\documentclass[sigconf]{acmart}
\usepackage{algorithm}
\usepackage{algorithmic}
\usepackage{multirow}
\PassOptionsToPackage{table,xcdraw,dvipsnames}{xcolor}
\usepackage[normalem]{ulem}
\usepackage{colortbl}
\usepackage{arydshln}
\usepackage{adjustbox}
\usepackage{pifont}

\definecolor{SteelBlue}{RGB}{70,130,180}
\definecolor{IndianRed}{RGB}{205,92,92}

\AtBeginDocument{%
  }

\setcopyright{acmlicensed}
\copyrightyear{2025}
\acmYear{2025}
\acmDOI{XXXXXXX.XXXXXXX}
\acmConference[MM '25]{the 33rd ACM International Conference on Multimedia}{October 27--31,
  2025}{Dublin, Ireland}
\acmISBN{978-1-4503-XXXX-X/2018/06}

\acmSubmissionID{2935}

\begin{document}

\title{MST-Distill: Mixture of Specialized Teachers for Cross-Modal Knowledge Distillation}

\author{Hui Li}
\email{gray1y@stu.xidian.edu.cn}
\orcid{0000-0003-4001-1161}
\affiliation{%
	\institution{Xidian University}
	\city{Xi'an}
	\country{China}
}

\author{Pengfei Yang}
\authornote{Corresponding author.}
\email{pfyang@xidian.edu.cn}
\affiliation{%
	\institution{Xidian University}
	\city{Xi'an}
	\country{China}
}

\author{Juanyang Chen}
\email{jychen\_324@stu.xidian.edu.cn}
\orcid{0009-0004-6143-5339}
\affiliation{%
	\institution{Xidian University}
	\city{Xi'an}
	\country{China}
}

\author{Le Dong}
\email{dongle@xidian.edu.cn}
\affiliation{%
	\institution{Xidian University}
	\city{Xi'an}
	\country{China}
}

\author{Yanxin Chen}
\email{24031110051@stu.xidian.edu.cn}
\orcid{0009-0001-5789-563X}
\affiliation{%
	\institution{Xidian University}
	\city{Xi'an}
	\country{China}
}

\author{Quan Wang}
\email{qwang@xidian.edu.cn}
\affiliation{%
	\institution{Xidian University}
	\city{Xi'an}
	\country{China}
}

\renewcommand{\shortauthors}{Li et al.}

\begin{abstract}
Knowledge distillation as an efficient knowledge transfer technique, has achieved remarkable success in unimodal scenarios. However, in cross-modal settings, conventional distillation methods encounter significant challenges due to data and statistical heterogeneities, failing to leverage the complementary prior knowledge embedded in cross-modal teacher models. This paper empirically reveals two critical issues in existing approaches: distillation path selection and knowledge drift. To address these limitations, we propose MST-Distill, a novel cross-modal knowledge distillation framework featuring a mixture of specialized teachers. Our approach employs a diverse ensemble of teacher models across both cross-modal and multimodal configurations, integrated with an instance-level routing network that facilitates adaptive and dynamic distillation. This architecture effectively transcends the constraints of traditional methods that rely on monotonous and static teacher models. Additionally, we introduce a plug-in masking module, independently trained to suppress modality-specific discrepancies and reconstruct teacher representations, thereby mitigating knowledge drift and enhancing transfer effectiveness. Extensive experiments across five diverse multimodal datasets, spanning visual, audio, and text, demonstrate that our method significantly outperforms existing state-of-the-art knowledge distillation methods in cross-modal distillation tasks. The source code is available at \url{https://github.com/Gray-OREO/MST-Distill}.
\end{abstract}





\maketitle

\begin{figure}[htbp]
	\centering
	\includegraphics[width=\linewidth]{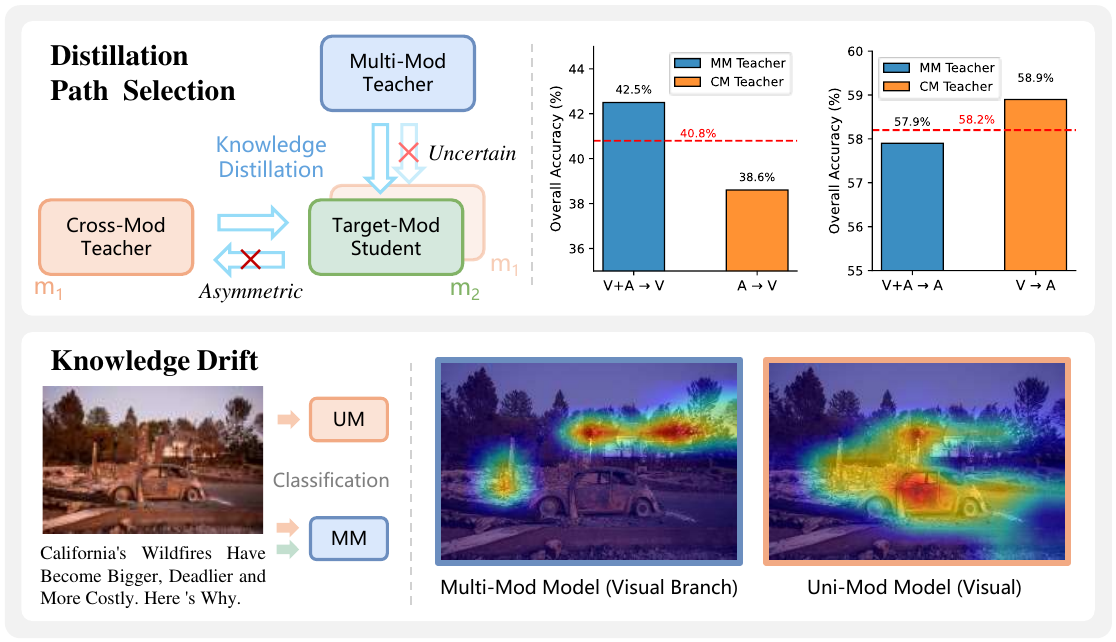}
	\caption{Illustration of two key challenges in cross-modal knowledge distillation: distillation path selection and knowledge drift. Top: Performance comparison of unimodal students under different teachers (multimodal \& cross-modal) on VGGSound-50k (visual \& audio), with red lines as baselines. Bottom: Grad-CAM comparison between multimodal and cross-modal teachers on CrisisMMD-V2 (visual \& text).}
	\label{fig:motivation}
\end{figure}

\section{Introduction}
With the rapid advancement of sensor technologies and intelligent devices, data acquisition methods have diversified significantly, generating abundant multimodal data across vision, audio, and text domains. This multimodal data provides rich training resources \cite{ref1} and offers a more comprehensive perspective \cite{ref2} for artificial intelligence models. By leveraging complementary cross-modal information, multimodal approaches have achieved remarkable performance in video understanding \cite{ref3}, cross-modal retrieval \cite{ref4}, and human-computer interaction \cite{ref5}.

The expansion of these multimodal applications has driven an increasing trend toward distributed and near-sensor computing paradigms. Edge computing, which deploys intelligent systems closer to data sources, offers substantial benefits including reduced latency, enhanced privacy, and decreased bandwidth requirements, making it ideal for time-sensitive and data-intensive applications \cite{ecc}. However, these edge deployments face significant challenges from dynamic network conditions and heterogeneous sensor characteristics \cite{ref6, ref7}. While conventional multimodal fusion methods assume well-aligned cross-modal correlations \cite{ref9}, real-world scenarios often involve temporal misalignment and partial modality absence due to asynchronous data transmission, degrading performance \cite{ref6}. In this context, cross-modal knowledge transfer emerges as a promising solution for resource-constrained edge devices, enabling efficient inference by constructing shared semantic spaces that facilitate the transfer of complementary knowledge across modalities.

Knowledge distillation (KD) offers an effective technique for model compression and knowledge transfer within a teacher-student framework \cite{kd}. Based on their supervision mechanisms, distillation methods are categorized as response-based \cite{kd, fitnets, rkd}, feature-based \cite{fitnets, ref15, fsp}, and relation-based \cite{rkd, crd}. While these approaches have demonstrated success in computer vision \cite{od} and natural language processing \cite{ref20}, they primarily address knowledge transfer within a single modality. When applied to cross-modal scenarios, these methods encounter additional challenges from data and statistical heterogeneity \cite{c2kd}, which violates the distributional consistency assumption \cite{kd_insights} in traditional distillation approaches, resulting in misaligned representations \cite{mfh} and unreliable knowledge transfer \cite{c2kd}. Building on these foundational challenges, we empirically identify two critical issues that remain underexplored in existing literature:

(1) \textbf{Path selection in cross-modal knowledge distillation.} As illustrated in Figure~\ref{fig:motivation} (top), modality imbalance \cite{c2kd} manifest as pervasive asymmetry and uncertainty in the process of cross-modal knowledge distillation. Knowledge from certain source modalities exhibits stronger transferability for specific tasks, while the reverse direction often yields poor performance. Moreover, even multimodal teachers, despite their potential to leverage complementary information across modalities, do not always provide effective supervisory signals, introducing significant challenges to designing efficient and generalizable distillation paradigms.

(2) \textbf{Knowledge drift between teacher and student models.} A fundamental challenge in cross-modal knowledge distillation stems from inductive bias mismatches between models trained on different data domains. Even multimodal teachers may exhibit unimodal bias \cite{unim_bias}, resulting in substantial discrepancies between the teacher's attention regions and those of the unimodal student when processing identical inputs. As shown in Figure~\ref{fig:motivation} (bottom), the Grad-CAM \cite{grad_cam} visualizations clearly illustrate these differences, revealing a knowledge drift that impacts model behavior and transfer effectiveness. We provide comprehensive empirical analysis and in-depth discussion of both challenges in Appendix~\ref{appendix:moti}.

To tackle the challenges of distillation path selection and knowledge drift in cross-modal knowledge distillation, we propose MST-Distill, a generalized and adaptive framework. By integrating diverse teacher models with an instance-level routing network, MST-Distill enables the target-modality student to dynamically select optimal distillation paths during training, thereby facilitating robust and flexible knowledge transfer. In addition, we introduce a plug-in MaskNet module that reconstructs teacher representations under the guidance of response consistency, encouraging behavioral alignment and mitigating knowledge drift across modalities. Our main contributions are summarized as follows:

\begin{itemize}
\item We point out two key challenges in cross-modal knowledge distillation: distillation path selection and knowledge drift. To address them, we propose MST-Distill, a unified framework integrating instance-level dynamic routing with reconstruction-consistency-guided teacher specialization mechanisms.

\item We construct a mixture of teachers comprising both multimodal and cross-modal models, coupled with an instance-level routing network that allows the student model to adaptability select the optimal distillation path.

\item To mitigate knowledge drift caused by inductive bias discrepancies between the models, we introduce a learnable MaskNet module that effectively suppresses modality-specific discrepancies while reconstructing teacher representations aligned with the student's behavior.

\item Extensive experiments across five datasets demonstrate the effectiveness and generalizability of MST-Distill in cross-modal knowledge distillation tasks.
\end{itemize}

\section{Related Work}

\subsection{Multimodal Learning}
Multimodal learning has emerged as a prominent research focus in computer vision \cite{ref24} and natural language processing \cite{ref27}. By integrating heterogeneous modality information, these approaches enable more comprehensive representation learning \cite{ref8, ref28} and have demonstrated efficacy in applications including sentiment analysis \cite{ref30}, video understanding \cite{ref31}, and multimodal dialogue systems \cite{ref33}.

However, multimodal learning inherently faces significant challenges due to its complexity, spanning network architecture design \cite{ref34}, cross-modal distributional differences \cite{ref37}, and optimization strategies \cite{mmpareto}. During training, these complexities manifest primarily as two fundamental obstacles that impede effective knowledge integration: modality conflict \cite{ref38} and unimodal bias \cite{ref39, ref40}. Modality conflict emerges when semantic and structural inconsistencies between modalities destabilize the optimization process, while unimodal bias \cite{unim_bias} occurs when training dynamics favor dominant modalities, suppressing information from others. These training challenges represent distinct barriers to modeling effective cross-modal relationships, ultimately constraining performance on downstream tasks. Building upon these foundational studies, researchers have made significant advances in addressing multimodal learning challenges. Zhang et al. \cite{unim_bias} revealed critical architectural limitations in late fusion models that promote unimodal bias, while Wei et al. \cite{mmpareto} introduced MMPareto, an optimization framework that effectively mitigates early-stage gradient conflicts through a dynamic systems approach. Complementing these efforts, Fan et al. \cite{ref43} developed a prototypical modal rebalancing method that strategically applies task-oriented unimodal constraints to counteract modality imbalance.

Although these contributions provide valuable theoretical insights and practical strategies for improving multimodal architectures and training methodologies, significant challenges persist. Particularly, the intrinsic difficulty of balancing modality-specific knowledge within multimodal frameworks continues to limit the effectiveness of cross-modal knowledge distillation, with knowledge drift emerging as a critical barrier to achieving optimal performance gains.

\subsection{Cross-Modal Knowledge Distillation}
Knowledge distillation proves highly effective for unimodal knowledge transfer, yet cross-modal scenarios present unique challenges from heterogeneous data formats and statistical discrepancies \cite{ref9}. Cross-Modal Knowledge Distillation (CMKD) addresses these issues by specifically bridging the modality gap, enhancing both representation quality and performance in target modalities through effective cross-modal knowledge alignment.

Early CMKD research focused on knowledge transfer between visually similar modalities (RGB, depth, infrared) that share visual characteristics despite different sensing principles \cite{nmmm}. The field has since expanded to highly heterogeneous modality pairs including vision, audio, and text \cite{cmkd_app1}, driving interest in modality complementarity and collaborative learning. To address structural and semantic discrepancies between diverse modalities, recent work has introduced sophisticated strategies such as contrastive learning \cite{vggs-50k}, modality decoupling \cite{mdecopling}, shared semantic representations \cite{cmkd_ssr}, and meta-learning \cite{metalearning}. These approaches have shown considerable efficacy across multimodal applications including video understanding, emotion recognition, and cross-modal retrieval.

Despite these advances, current CMKD approaches remain primarily constrained to specific scenarios and distillation configurations, limiting their adaptability across diverse modality combinations and task requirements. Addressing this constraint, Xue et al. \cite{mke} introduced MKE, demonstrating viable cross-modal knowledge exchange between unimodal and multimodal models under generalized conditions. They subsequently developed the modality focusing hypothesis \cite{mfh}, approaching CMKD through strategic construction and shaping of the teacher model's feature space. In a recent contribution, Huo et al. \cite{c2kd} proposed C\textsuperscript{2}KD, an innovative framework leveraging soft label rank consistency to guide dynamic sample selection for optimized knowledge transfer.

While these efforts have established robust theoretical and optimization foundations for CMKD, significant opportunities remain in leveraging diverse teacher models and developing learnable mechanisms for behavioral alignment. To address these opportunities, we propose MST-Distill, a generalized framework for cross-modal knowledge distillation that systematically addresses distillation path selection and behavioral alignment between models. This approach enables robust and adaptive knowledge transfer across heterogeneous modalities, advancing the state-of-the-art in cross-modal knowledge distillation.

\begin{figure*}[htbp]
	\centering
	\includegraphics[width=\linewidth]{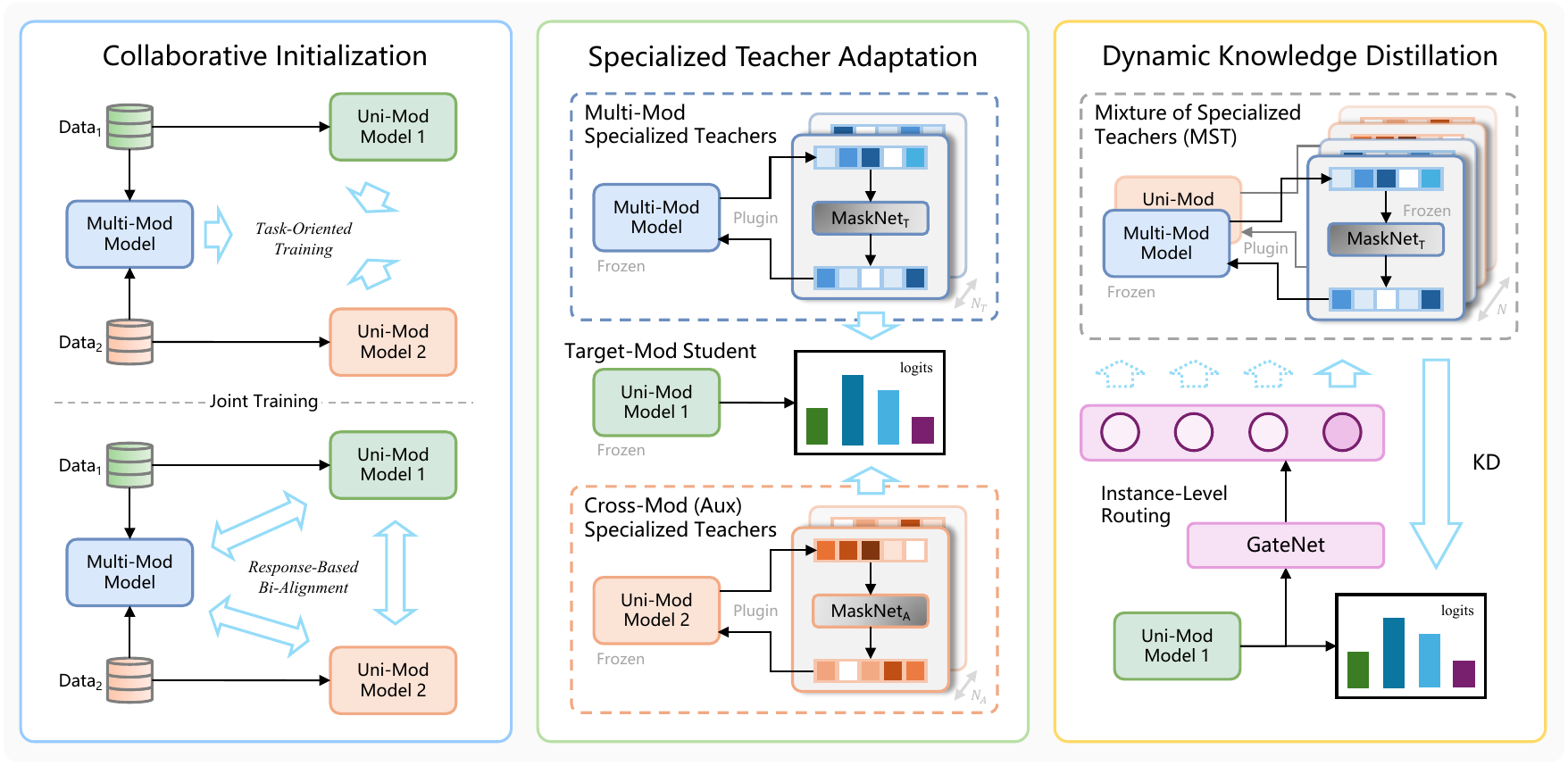}
	\caption{Overview of the MST-Distill framework in a two-modality setting, consisting of three stages: Collaborative Initialization (CI), Specialized Teacher Adaptation (STA), and Dynamic Knowledge Distillation (DKD).}
	\label{fig:framework}
\end{figure*}

\section{Method}
In this section, we introduce the detailed implementation of our proposed Mixture of Specialized Teachers for Cross-Modal Knowledge Distillation (MST-Distill). An overview of MST-Distill framework is illustrated in Figure~\ref{fig:framework}, which is composed of three sequential stages: Collaborative Initialization (S1), Specialized Teacher Adaptation (S2), and Dynamic Knowledge Distillation (S3). In the following subsections, we elaborate on the objectives, mechanisms, and technical implementations of each stage.

\subsection{Collaborative Initialization of Modality-Specific Members}
As a further step toward exploring the potential benefits of teacher diversity in cross-modal knowledge transfer, introducing a set of diverse teacher models becomes a natural and effective design choice. We begin by formally defining the cross-modal knowledge distillation task. Let $\mathcal{D} = \{ ( x_1^{(s)}, x_2^{(s)}, \dots, x_M^{(s)}; y^{(s)} ) \}_{s=1}^{S}$ denote a multimodal dataset with $S$ samples, where each sample consists of $M$ data from different modalities and a corresponding label. The $i$-th modality is denoted by $m_i$, with $i=0$ indicating the multimodal case, where $x_0 = (x_1, x_2, \dots, x_M)$ combines all modality data for joint inference. The network model corresponding to modality $m_i$ is denoted as $f_{m_i}$.

In the first stage of MST-Distill, we do not specify a target student modality in advance. Instead, we treat all $M+1$ models equally as modality-specific members and train them jointly for the collaborative initialization. The training objective consists of two components: a task loss $\mathcal{\ell}_{task}$, which supervises all members using the ground-truth label, and an alignment loss $\mathcal{\ell}_{align}$, which encourages prediction consistency via bidirectional Kullback–Leibler (KL) divergence between all modality pairs.

For a single training sample, the losses are defined as:
\begin{equation}
\mathcal{\ell}_{\mathrm{task}} = \sum_{i=0}^{M} \mathrm{CE}\left ( f_{m_{i}} \left ( x_{i}; \theta _{m_{i}} \right ), y \right),
\end{equation}

\begin{equation}
\mathcal{\ell}_{\mathrm{align}} = \sum_{0 \le i < j \le M} \left[ \mathrm{KL} \left (P_{m_{i}} \parallel P_{m_{j}}\right) + \mathrm{KL} \left(P_{m_{j}} \parallel P_{m_{i}}\right) \right],
\end{equation}

\begin{equation}
P_{m_i} = \text{softmax}\left( f_{m_i}\left((x_i; \theta_{m_i}\right) / \tau \right), \quad i \in \{0, \dots, M\},
\end{equation}
where $\mathrm{CE} ( \cdot )$ denotes the cross-entropy loss, $\mathrm{KL} ( \cdot )$ is the Kullback-Leibler divergence, $\theta_{m_i}$ is the parameter set of model $f_{m_i}$, and $P_{m_i}$ is the softened output distribution with temperature $\tau$.

Notably, we do not apply gradient detachment to the outputs of the teacher models, in contrast to conventional bidirectional distillation practices, which enables mutual gradient propagation among modality-specific members. The loss function in Stage 1 for a minibatch of size $B$ is given by:
\begin{equation}
	\label{eq:s1}
	\mathcal{L}_{\mathrm{S_{1}}} = \frac{1}{B} \sum_{b=1}^{B} \left ( \mathcal{\ell}_{\mathrm{task}}^{\left ( b \right ) }+\mathcal{\ell}_{\mathrm{align}}^{\left ( b \right ) }\right ),
\end{equation}
where $\mathcal{\ell}^{\left ( b \right )}$ represents the per-sample loss corresponding to the $b$-th instance in the minibatch.

\begin{figure}[htbp]
	\centering
	\includegraphics[width=\linewidth]{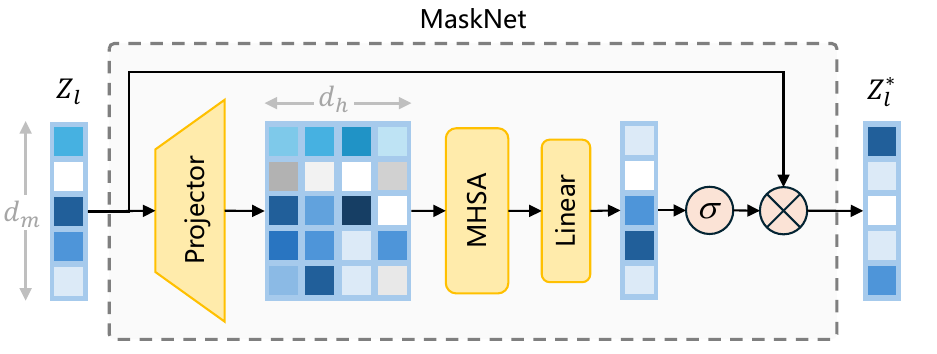}
	\caption{Overall architecture of MaskNet. A soft mask is generated to reconstruct the intermediate feature maps of the teacher model through a standard multi-head self-attention mechanism.}
	\label{fig:mn}
\end{figure}

\subsection{MaskNet-Driven Specialized Teacher Adaptation}
Inspired by the feature significance-based filtering strategy in \cite{mfh}, which suppresses non-salient features in teacher representations, we introduce a learnable, plug-in module called MaskNet in the second stage of MST-Distill. This module generalizes soft masking-based reconstruction to arbitrary intermediate layers of teacher models, thereby enabling efficient behavioral alignment between teachers and the target student model.

As illustrated in the STA module of Figure~\ref{fig:framework}, the selection of the target modality $m_t$ determines the corresponding unimodal student, the multimodal teacher, and the auxiliary cross-modal teachers. To further enhance the diversity of the teacher ensemble, we insert independently parameterized MaskNet modules into selected intermediate layers of each teacher model. Specifically, for a single base teacher model, we create multiple specialized versions by incorporating different MaskNet instances at the same intermediate layers, where each MaskNet has its own independent parameters. This approach allows us to derive $N$ specialized teacher while reusing the base architecture, calculated as:
\begin{equation}
N = N_{T,m_{0}} + \sum_{\substack{i=1 \\ i \ne t}}^{M} N_{A,m_{i}} = \sum_{\substack{i=0 \\ i \ne t}}^{M} N_{m_{i}},
\end{equation}
where $N_{T,m_{0}}$ and $N_{A,m_{i}}$ denote the number of selected reconstruction layers in the multimodal teacher and auxiliary teachers, respectively. To simplify notation, we uniformly denote the number of selected layers from each teacher as $N_{m_i}$.

The structure of MaskNet is illustrated in Figure~\ref{fig:mn}. For a given intermediate feature $Z_l \in \mathbb{R}^{d_{m}}$ from the $l$-th layer of a teacher model, MaskNet first projects it into a latent space $\mathbb{R}^{d_{m}\times d_{h}}$ via the projector (a linear layer followed by reshaping), and $d_{m}$, $d_{h}$ are the input and hidden dimensions, respectively. This is followed by a multi-head self-attention (MHSA) block \cite{transformer} and a linear layer, with a sigmoid activation to produce a soft attention mask. The masked feature $Z_l^{*} \in \mathbb{R}^{d_{m}}$ is then obtained by an element-wise Hadamard product between the input and the soft mask. The process is formally defined as:
\begin{align}
	Z_{l}^{*}&= \mathrm{MaskNet}\left ( Z_{l};\theta _{\mathrm{MN}} \right ) \notag \\
	&= \sigma \left (\mathrm{Linear}\left ( \mathrm{MHSA} \left (\mathrm{Projector}\left ( Z_{l}\right )\right )\right )\right )\otimes Z_{l},
\end{align}
where $\theta _{\mathrm{MN}}$ denotes the parameters of the MaskNet module, and $\otimes$ represents the Hadamard product operator.

Subsequently, all model parameters except those of MaskNet are frozen, and each MaskNet is trained independently to align the behavior of its corresponding specialized teacher with that of the target student, guided by response consistency. Specifically, let $f_{m_{\delta(j)}}^j$ denote the $j$-th specialized teacher model with a corresponding MaskNet module, where $\delta(j)$ is an index mapping function that identifies the modality source of the $j$-th teacher:
\begin{equation}
\delta(j) = \arg\min_{\substack{i\in \left \{ 0,\dots ,M \right \} \\ i \ne t }} \left\{ j \le \sum_{k=0}^{i} N_{m_{k}} \right\}, \quad j\in \left \{ 1,\dots , N  \right \},
\end{equation}
where $N_{m_{k}}$ is the number of specialized teachers derived from modality $m_{k}$.

For each training sample, the discrepancy between the output distributions of the $j$-th teacher and the target-modality student is measured as:
\begin{equation}
\mathcal{\ell}_{j}= \mathrm{KL}\left(Q_{m_{t}}^{j} \parallel Q_{m_{\delta(j)}}^{j}\right),
\end{equation}

\begin{equation}
Q_{m_i}^{j} = \text{softmax}\left( f_{m_i}^{j}\left(x_i; \theta_{m_i}, \theta_{\mathrm{MN}}^{j}\right) / \tau \right), \quad i \in \{0, \dots, M\},
\end{equation}
where $f_{m_i}^{j}$ is the $j$-th specialized teacher under modality $m_i$,  $Q_{m_i}^{j}$ denotes its corresponding softened output distribution, and $\theta_{\mathrm{MN}}^{j}$ is the associated MaskNet parameter. As a result, the loss function in Stage 2 for the $j$-th teacher over a minibatch of size $B$ is defined as:
\begin{equation}
\label{eq:s2}
\mathcal{L}_{\mathrm{S_{2}}}^{j} = \frac{1}{B}  \sum_{b=1}^{B} \mathcal{\ell}_{j}^{\left ( b \right ) }.
\end{equation}
\subsection{Dynamic Knowledge Distillation with a Mixture of Specialized Teachers}
As the final stage of MST-Distill, the dynamic knowledge distillation process focuses on adaptively selecting and leveraging specialized teachers for the target modality $m_t$. Given an input data $x_t$ from modality $m_t$, the corresponding student model $f_{m_t}$ produces a logits vector $Z_{out}$, which is then passed into a routing network (GateNet) to generate confidence scores $C\in \mathbb{R} ^{N}$ over all specialized teachers:
\begin{equation}
Z_{out} = f_{m_t}\left(x_t; \theta_{m_t}\right), \quad t \in \left\{ 1, \dots, M \right\},
\end{equation}

\begin{equation}
C = \text{softmax} \left( \mathrm{GateNet}\left(Z_{out};\theta _{\mathrm{GN}}\right ) \right ),
\end{equation}
where $\mathrm{GateNet(\cdot)}$ is the routing network with parameters $\theta _{\mathrm{GN}}$, implemented as a multi-layer perceptron (MLP) with $N$ output nodes.

Based on the confidence scores $C$, we adopt the TopK rule to select the indices of the $k$ highest-scoring teachers in descending order of confidence:
\begin{equation}
\mathcal{T}_{\text{top-}k}=\mathrm{TopK}_{k}\left ( \left \{ C^{j} \right \}^{N}_{j=1}  \right ),
\end{equation}
where $\mathcal{T}_{\text{top-}k}$ denotes the indices of the selected teachers used for subsequent knowledge distillation.

Notably, this instance-wise selection allows the student to dynamically choose the teachers with the highest cross-modal transferability. Given the selected top-$k$ teachers, the distillation loss for each training sample is computed by measuring the KL divergence between the softened outputs of the student and those of the selected teachers:
\begin{equation}
\mathcal{\ell}_{DKD} = \sum_{j\in \mathcal{T}_{\text{top-}k}} \mathrm{KL}\left(Q_{m_{\delta(j) }}^{j} \parallel P_{m_t}\right),
\end{equation}
where $Q_{m_{\delta(j) }}^{j}$ denotes the softened output distribution from the $j$-th specialized teacher of modality $m_{\delta(j) }$, and $P_{m_t}$ is the student’s output distribution.

Similar to \cite{kd}, the task-specific classification loss for the student is computed as:
\begin{equation}
\mathcal{\ell}_{S} = \mathrm{CE}\left ( f_{t} \left ( x_{t}; \theta _{m_{t}} \right ), y \right).
\end{equation}

Furthermore, to prevent the routing network from converging to only a small subset of teachers, we incorporate a load balancing loss that promotes diverse teacher utilization.

Specifically, for each mini-batch, we calculate the average confidence distribution across all samples and compare it to a uniform distribution $U\in \mathbb{R} ^{N}$ using the Kullback-Leibler divergence:
\begin{equation}
\bar{C} = \frac{1}{B} \sum_{b=1}^{B} C^{\left ( b \right ) },
\end{equation}

\begin{equation}
\mathcal{L}_{LB} = \mathrm{KL}\left(U \parallel \bar{C} \right),
\end{equation}
where each element in $U$ is set to $\frac{1}{N}$.

This encourages uniform utilization of all specialized teachers throughout training. And the final loss function for Stage 3 over a mini-batch is defined as:

\begin{equation}
\label{eq:s3}
\mathcal{L}_{\mathrm{S_{3}}} = \frac{1}{B} \sum_{b=1}^{B}\left ( \mathcal{\ell}_{S}^{\left ( b \right ) } + \lambda_{1} \cdot \mathcal{\ell}_{DKD}^{\left ( b \right ) }  \right ) + \lambda_{2} \cdot \mathcal{L}_{LB},
\end{equation}
where $\lambda_{1}$ and $\lambda_{2}$ are decay-weighted hyperparameters that gradually decrease during training. See Appendix~\ref{appendix:mst_pseudocode} for the detailed pseudocode of MST-Distill.

\renewcommand{\arraystretch}{0.7}
\begin{table*}[t]
	\centering
	\caption{Performance comparison on multimodal classification datasets. The evaluation metric is the average overall accuracy over five independent runs. Dashed lines separate baselines trained independently on each modality. CM and MM denote cross-modal and multimodal teachers, respectively. Distillation results with performance gains are highlighted in \textcolor{SteelBlue}{Blue}. The top two performing results are shown in \textcolor{IndianRed}{Red}, with the best result further \textcolor{IndianRed}{\uline{underlined}} for emphasis.}
	\label{tab:performance_compar_cls}
	\begin{tabular}{ccccccccccc}
		\toprule
		\multirow{2}{*}{\textbf{Paradigm}} & \multirow{2}{*}{\textbf{Method}} & \multirow{2}{*}{\textbf{T-Config}} & \multicolumn{2}{c}{\textbf{AV-MNIST}} & \multicolumn{2}{c}{\textbf{RAVDESS}} & \multicolumn{2}{c}{\textbf{VGGSound-50k}} & \multicolumn{2}{c}{\textbf{CrisisMMD-V2}} \\
		& & & Image & Audio & Visual & Audio & Visual & Audio & Image & Text \\
		\midrule
		\textemdash & w/o KD & \textemdash & 0.6322 & 0.4372 & 0.7472 & 0.6694 & 0.4080 & 0.5818 & 0.5524 & 0.5405 \\
		\hdashline
		\noalign{\vskip 2pt}
		\multirow{4}{*}{Response-Based} 
		& \multirow{2}{*}{KD} & MM & 0.6322 & {\color{SteelBlue} 0.4383} & {\color{SteelBlue} 0.7521} & {\color{SteelBlue} 0.6979} & {\color{SteelBlue} 0.4248} & 0.5786 & 0.5505 & 0.5367 \\
		&                      & CM & {\color{SteelBlue} 0.6328} & {\color{SteelBlue} 0.4377} & 0.7035 & {\color{SteelBlue} 0.6944} & 0.3855 & {\color{SteelBlue} 0.5887} & 0.5501 & 0.5377 \\
		\arrayrulecolor{lightgray}\cmidrule(lr){2-11}\arrayrulecolor{black}
		& \multirow{2}{*}{MLLD} & MM & 0.6288 & 0.4372 & {\color{SteelBlue} 0.7549} & {\color{SteelBlue} 0.6930} & {\color{SteelBlue} 0.4241} & {\color{SteelBlue} 0.5868} & {\color{IndianRed} 0.5549} & 0.5098 \\
		&                      & CM & 0.6327 & {\color{SteelBlue} 0.4380} & 0.7208 & {\color{SteelBlue} 0.6799} & 0.4044 & {\color{SteelBlue} 0.5898} & {\color{SteelBlue} 0.5541} & 0.5127 \\
		\midrule
		\multirow{4}{*}{Feature-Based} 
		& \multirow{2}{*}{FitNets} & MM & 0.6255 & 0.4332 & 0.7382 & 0.6438 & 0.3778 & 0.5598 & {\color{SteelBlue} 0.5538} & 0.5317 \\
		&                      & CM & 0.6201 & 0.4292 & 0.6889 & 0.6229 & 0.3742 & 0.5685 & 0.5458 & 0.5165 \\
		\arrayrulecolor{lightgray}\cmidrule(lr){2-11}\arrayrulecolor{black}
		& \multirow{2}{*}{OFA} & MM & {\color{SteelBlue} 0.6331} & 0.4367 & 0.7333 & 0.6570 & 0.3808 & {\color{SteelBlue} 0.5821} & {\color{SteelBlue} 0.5509} & 0.5394 \\
		&                      & CM & {\color{SteelBlue} 0.6334} & {\color{SteelBlue} 0.4380} & 0.7292 & 0.6445 & 0.3861 & {\color{SteelBlue} 0.5839} & 0.5493 & {\color{IndianRed} 0.5420} \\
		\midrule
		\multirow{4}{*}{Relation-Based} 
		& \multirow{2}{*}{RKD} & MM & {\color{IndianRed} 0.6341} & 0.3868 & {\color{SteelBlue} 0.7486} & {\color{SteelBlue} 0.6792} & {\color{SteelBlue} 0.4156} & 0.5629 & 0.5499 & {\color{SteelBlue} 0.5406} \\
		&                      & CM & {\color{SteelBlue} 0.6337} & 0.3850 & {\color{SteelBlue} 0.7569} & {\color{SteelBlue} 0.6826} & {\color{SteelBlue} 0.4092} & 0.5709 & 0.5508 & {\color{SteelBlue} 0.5419} \\
		\arrayrulecolor{lightgray}\cmidrule(lr){2-11}\arrayrulecolor{black}
		& \multirow{2}{*}{CRD} & MM & {\color{SteelBlue} 0.6333} & {\color{SteelBlue} 0.4387} & {\color{SteelBlue} 0.7681} & {\color{SteelBlue} 0.6993} & 0.3804 & {\color{SteelBlue} 0.5826} & {\color{IndianRed} \uline{0.5559}} & 0.5268 \\
		&                      & CM & 0.6281 & {\color{IndianRed} 0.4389} & {\color{SteelBlue} 0.7736} & {\color{SteelBlue} 0.7014} & 0.3537 & {\color{SteelBlue} 0.5826} & 0.5504 & 0.5256 \\
		\midrule
		Mutual-Learning & DML & MM+CM & 0.6320 & {\color{IndianRed} \uline{0.4393}} & {\color{SteelBlue} 0.7660} & {\color{IndianRed} \uline{0.7202}} & {\color{IndianRed} \uline{0.4601}} & {\color{SteelBlue} 0.5915} & 0.5465 & 0.5351 \\
		\midrule
		\multirow{3}{*}{Cross-Modal} 
		& MGDFR & CM & {\color{SteelBlue} 0.6329} & {\color{IndianRed} 0.4389} & {\color{SteelBlue} 0.7576} & {\color{SteelBlue} 0.7111} & {\color{SteelBlue} 0.4375} & {\color{SteelBlue} 0.5967} & 0.5469 & 0.5390 \\
		& C\textsuperscript{2}KD
		& CM & 0.6309 & {\color{IndianRed} 0.4389} & {\color{IndianRed} 0.7754} & 0.6500 & {\color{SteelBlue} 0.4229} & {\color{IndianRed} 0.6080} & {\color{SteelBlue} 0.5528} & 0.5204 \\
		\rowcolor{gray!10}
		& \textbf{MST-Distill} & MM+CM & {\color{IndianRed} \uline{0.6359}} & {\color{SteelBlue} 0.4381} & {\color{IndianRed} \uline{0.7868}} & {\color{IndianRed} 0.7174} & {\color{IndianRed} 0.4595} & {\color{IndianRed} \uline{0.6098}} & 0.5495 & {\color{IndianRed} \uline{0.5466}} \\
		\bottomrule
	\end{tabular}
\end{table*}


\section{Experiments}
We conduct extensive experiments on five datasets encompassing diverse modality combinations, covering both cross-modal classification and semantic segmentation tasks. To comprehensively assess the effectiveness of our approach, we compare it with a broad range of representative baselines, including response-based KD \cite{kd} and MLLD \cite{mlld}, feature-based FitNets \cite{fitnets} and OFA \cite{ofa}, relation-based RKD \cite{rkd} and CRD \cite{crd}, the mutual learning method DML \cite{dml}, and two cross-modal knowledge distillation approaches, MGDFR (reproduced from \cite{mfh}) and C\textsuperscript{2}KD \cite{c2kd}. Comprehensive ablation and sensitivity studies are further conducted to gain deeper insights into the effectiveness of the proposed MST-Distill framework.

\subsection{Multimodal Classification}
\label{sec:mmcls}
We follow \cite{mfh, c2kd} and conduct experiments on four visual-audio and image-text datasets: (1) \textbf{AV-MNIST} \cite{avmnist} is a visual–audio dataset for digit classification, covering 10 categories of paired handwritten digits and spoken audio spectrograms. (2) \textbf{RAVDESS} \cite{ravdess} is a visual–audio dataset for emotion recognition, with 8 emotional categories expressed through aligned facial and vocal cues. (3) \textbf{VGGSound-50k} \cite{vggs} is a visual–audio scene classification dataset spanning 141 real-world categories \cite{vggs-50k} with co-occurring sound and visual content. (4) \textbf{CrisisMMD-V2} \cite{cmmdv2} is an image–text dataset for humanitarian classification, comprising 8 crisis-related categories based on image–text pairs from social media. Further details of these datasets are provided in Appendix~\ref{appendix:dataset}.

\paragraph{Implementation.} We adopt a consistent preprocessing strategy following \cite{mfh,c2kd}. For each dataset, we use customized but consistent teacher and student architectures across all methods, detailed in Appendix~\ref{appendix:model}. All distillation methods use identical training conditions: 100 epochs for single-stage methods, consistent sub-stage epochs for multi-stage methods (FitNets, MGDFR, and ours), with uniform batch size and loss decay schedules. In our method, MaskNets with three self-attention heads are inserted into intermediate and penultimate layers of teacher networks (using post-fusion features for multimodal teachers). We implement Top-$k$ dynamic distillation with $k=1$, setting initial $\lambda_1$ and $\lambda_2$ to 1, with decay schedules of halving every 30 epochs and 10\% reduction every 10 epochs, respectively. Data is split 60\%/20\%/20\% for training/validation/testing, and results are averaged over five runs using the best validation model. All experiments are conducted on a server equipped with an Intel Xeon Gold 6248R CPU and an NVIDIA A100 GPU.

\paragraph{Comparison Results.} We compare our MST-Distill framework against several advanced knowledge distillation baselines under identical training settings. As shown in Table~\ref{tab:performance_compar_cls}, our proposed framework MST-Distill achieves either the best or the second-best performance across all four multimodal datasets. The advantages are particularly evident on datasets with pronounced modality imbalance, such as AV-MNIST, RAVDESS, and VGGSound-50k, demonstrating the robustness and generalization capability of our method in diverse cross-modal scenarios. In comparison to a wide range of traditional knowledge distillation methods and recent cross-modal approaches, MST-Distill consistently delivers superior performance. It is worth noting that DML, a general-purpose mutual learning method originally designed for unimodal settings, also achieves competitive results in the cross-modal domain. This further confirms the importance of utilizing diverse and collaborative teacher signals to enhance knowledge transfer. Furthermore, we observe that feature-based methods including FitNets and OFA tend to underperform. This may be attributed to their reliance on feature-level similarity, which can be inadequate for capturing the complementary knowledge across heterogeneous modalities. In contrast, relation-based methods such as RKD and CRD exhibit better compatibility, as the structural relations among samples are relatively more stable across modalities, making them more suitable for cross-modal distillation tasks.

\subsection{Multimodal Semantic Segmentation}
We further evaluate the generalization of MST-Distill on the multimodal semantic segmentation task, focusing on knowledge transfer between closely related modalities (RGB and depth). Following the protocol in \cite{mfh}, we conduct experiments on the NYU-Depth-V2 dataset \cite{nyudv2}, which consists of 1,449 aligned RGB–depth image pairs with annotations for 40 semantic categories.

\renewcommand{\arraystretch}{0.7}
\begin{table}[htbp]
	\centering
	\caption{Performance comparison on NYU-Depth-V2 using Overall Accuracy (OA), Average Accuracy (AA), and mean IoU (mIoU) for both RGB and Depth modalities.}
	\label{tab:performance_compar_ss}
	\begin{adjustbox}{width=\linewidth}
		\begin{tabular}{cccccccc}
			\toprule
			\multirow{2}{*}{\textbf{Method}} & \multirow{2}{*}{\textbf{T-Config}} & \multicolumn{3}{c}{\textbf{RGB}} & \multicolumn{3}{c}{\textbf{Depth}} \\
			& & OA & AA & mIoU & OA & AA & mIoU \\
			\midrule
			w/o KD & \textemdash & 0.5189 & 0.2272 & 0.1453 & 0.5301 & 0.2015 & 0.1318 \\
			\hdashline
			\noalign{\vskip 2pt}
			\multirow{2}{*}{KD}
			& MM & \textcolor{SteelBlue}{0.5251} & \textcolor{SteelBlue}{0.2338} & \textcolor{SteelBlue}{0.1491} & \textcolor{SteelBlue}{0.5479} & \textcolor{IndianRed}{0.2493} & \textcolor{IndianRed}{0.1674} \\
			& CM & 0.5163 & 0.2202 & 0.1406 & \textcolor{SteelBlue}{0.5427} & \textcolor{SteelBlue}{0.2285} & \textcolor{SteelBlue}{0.1529} \\
			\arrayrulecolor{lightgray}\cmidrule(lr){1-8}\arrayrulecolor{black}
			\multirow{2}{*}{FitNets} 
			& MM & 0.5167 & 0.2077 & 0.1338 & \textcolor{SteelBlue}{0.5337} & 0.2006 & 0.1310 \\
			& CM & 0.4919 & 0.1636 & 0.1011 & \textcolor{SteelBlue}{0.5442} & \textcolor{SteelBlue}{0.2151} & \textcolor{SteelBlue}{0.1443} \\
			\arrayrulecolor{lightgray}\cmidrule(lr){1-8}\arrayrulecolor{black}
			DML & MM+CM & \textcolor{IndianRed}{\uline{0.5455}} & \textcolor{SteelBlue}{0.2402} & \textcolor{IndianRed}{0.1567} & \textcolor{IndianRed}{0.5570} & \textcolor{SteelBlue}{0.2357} & \textcolor{SteelBlue}{0.1590} \\
			MGDFR & CM & 0.5068 & 0.2004 & 0.1279 & \textcolor{SteelBlue}{0.5328} & \textcolor{SteelBlue}{0.2352} & \textcolor{SteelBlue}{0.1567} \\
			C\textsuperscript{2}KD
			& CM & \textcolor{SteelBlue}{0.5329} & \textcolor{SteelBlue}{0.2332} & \textcolor{SteelBlue}{0.1525} & \textcolor{SteelBlue}{0.5468} & \textcolor{SteelBlue}{0.2076} & \textcolor{SteelBlue}{0.1386} \\
			\rowcolor{gray!10}
			\textbf{MST-Distill} & MM+CM & \textcolor{IndianRed}{0.5396} & \textcolor{IndianRed}{\uline{0.2410}} & \textcolor{IndianRed}{\uline{0.1620}} & \textcolor{IndianRed}{\uline{0.5572}} & \textcolor{IndianRed}{\uline{0.2584}} & \textcolor{IndianRed}{\uline{0.1797}} \\
			\bottomrule
		\end{tabular}
	\end{adjustbox}
\end{table}

\paragraph{Implementation.} Consistent with our standardized setup in Section~\ref{sec:mmcls}, we employ FuseNet \cite{fusenet} as the multimodal teacher model and derive unimodal student models from its modality-specific branches. Unlike classification tasks that operate at the sample level, our distillation occurs at a finer pixel-wise granularity. To enhance feasibility, we apply knowledge transfer at the encoder-decoder bottleneck, focusing on mid-level features with lower dimensionality. Notably, due to the dense prediction nature of semantic segmentation, many traditional knowledge distillation methods designed for classification cannot be directly applied here. Thus, our comparison includes only a subset of representative response-based and feature-based methods.

\paragraph{Comparison Results.} To further validate the effectiveness of MST-Distill in transferring knowledge across closely related modalities for dense prediction tasks, we conduct multimodal semantic segmentation experiments on the NYU-Depth-V2 dataset. As shown in Table~\ref{tab:performance_compar_ss}, MST-Distill demonstrates superior performance across all evaluation metrics (OA, AA, and mIoU) for both RGB and depth modalities, ranking first in five of six metrics and second in the remaining one. Notably, MST-Distill attains the highest mIoU scores on both modalities, with 0.1620 for RGB and 0.1797 for depth, clearly outperforming all existing baseline methods. These improvements demonstrate the framework’s capability to effectively capture and transfer fine-grained, structured knowledge across similar modalities. Moreover, consistent with our classification results, the general-purpose distillation method DML also performs competitively in this task, further confirming the advantage of utilizing diverse teacher models for cross-modal knowledge transfer. In contrast, cross-modal methods such as C\textsuperscript{2}KD and MGDFR fall short of MST-Distill, partly due to the asymmetry in distillation effectiveness caused by reliance on a single cross-modal teacher. Overall, these results reinforce the adaptability and effectiveness of MST-Distill across both classification and segmentation tasks.

\renewcommand{\arraystretch}{0.8}
\begin{table}[htbp]
	\centering
	\small
	\caption{Ablation study of the three stages in MST-Distill. Each value indicates the mean accuracy averaged over all modality-specific students and five runs across four multimodal classification datasets.}
	\label{tab:ablation_main}
	\begin{adjustbox}{width=\linewidth}
		\begin{tabular}{
				l   
				c   
				c   
				c   
				*{4}{>{\centering\arraybackslash}m{1.5cm}}  
			}
			\toprule
			\textbf{Setting} & \textbf{S1} & \textbf{S2} & \textbf{S3} & \textbf{AV-MNIST} & \textbf{RAVDESS} & \textbf{VGGSound-50k} & \textbf{CrisisMMD-V2} \\
			\midrule
			baseline & \textemdash & \textemdash & \textemdash & 0.5347 & 0.7083 & 0.4949 & 0.5465 \\
			\hdashline
			\noalign{\vskip 2pt}
			(a) & ~ & ~ & ~ & 0.5362 & 0.7372 & 0.5273 & 0.5404 \\
			(b) & ~ & ~ & \checkmark & 0.5364 & 0.7372 & 0.5279 & 0.5381 \\
			(c) & \checkmark & ~ & \checkmark & 0.5369 & 0.7445 & \textcolor{IndianRed}{\uline{0.5406}} & 0.5393 \\
			(d) & ~ & \checkmark & \checkmark & 0.5362 & 0.7500 & 0.5271 & 0.5432 \\
			(e) & \checkmark & \checkmark & ~ & \textcolor{IndianRed}{\uline{0.5371}} & \textcolor{IndianRed}{0.7504} & \textcolor{IndianRed}{0.5349} & \textcolor{IndianRed}{0.5449} \\
			\rowcolor{gray!10}
			(f) \textbf{default} & \checkmark & \checkmark & \checkmark & \textcolor{IndianRed}{0.5370} & \textcolor{IndianRed}{\uline{0.7521}} & 0.5347 & \textcolor{IndianRed}{\uline{0.5481}} \\
			\bottomrule
		\end{tabular}
	\end{adjustbox}
\end{table}

\subsection{Ablation and Sensitivity Studies}
To comprehensively understand the effectiveness of the proposed framework, we conduct ablation and sensitivity studies from two perspectives. First, we perform systematic ablation experiments to evaluate the contribution of  the three core stages in MST-Distill. Following this, we investigate various configurations of the Mixture of Specialized Teachers (MST), such as modality diversity and the number of top-$k$ selected teachers, to analyze their impact on performance and robustness.

\subsubsection{Component Impacts of MST-Distill.}
Given the demonstrated advantages of diverse specialized teachers in prior experiments, we proceed with targeted ablation studies to investigate the specific contributions of each stage within the MST-Distill framework. Based on the established teacher diversity, we selectively activate three core stages: Collaborative Initialization (S1), Specialized Teacher Adaptation (S2), and Dynamic Knowledge Distillation (S3), in order to analyze their individual and combined impact. The results are shown in Table~\ref{tab:ablation_main}, which reveal three key observations:

(1) \textbf{Cross-modal knowledge distillation performs better with strongly aligned modality pairs.} Compared to the independently trained student (baseline), setting (a) shows that mean-based distillation from diverse teachers significantly improves performance on well-aligned tasks such as RAVDESS and VGGSound-50k, but provides limited or even negative effects on loosely aligned data such as AV-MNIST and CrisisMMD-V2.

(2) \textbf{The proposed dynamic knowledge distillation strategy relies heavily on early-stage collaborative training.} As shown in setting (b), applying dynamic distillation directly on static diverse teachers leads to minimal improvements. In contrast, both S1 (setting (c)) and S2 (setting (d)) introduce notable performance gains when compared to setting (b), highlighting the importance of model-teacher alignment and specialization before applying dynamic distillation. Interestingly, the benefits exhibit dataset-specific tendencies (RAVDESS benefits more from S2, while VGGSound-50k favors S1).

(3) \textbf{The transferability benefits from CI and STA are decoupled yet complementary.} Their combination in setting (e) brings further improvements over using either stage alone. Incorporating S3 in setting (f) achieves the best overall performance, confirming the effectiveness of the full three-stage framework.

\begin{figure}[t]
	\centering
	\includegraphics[width=\linewidth]{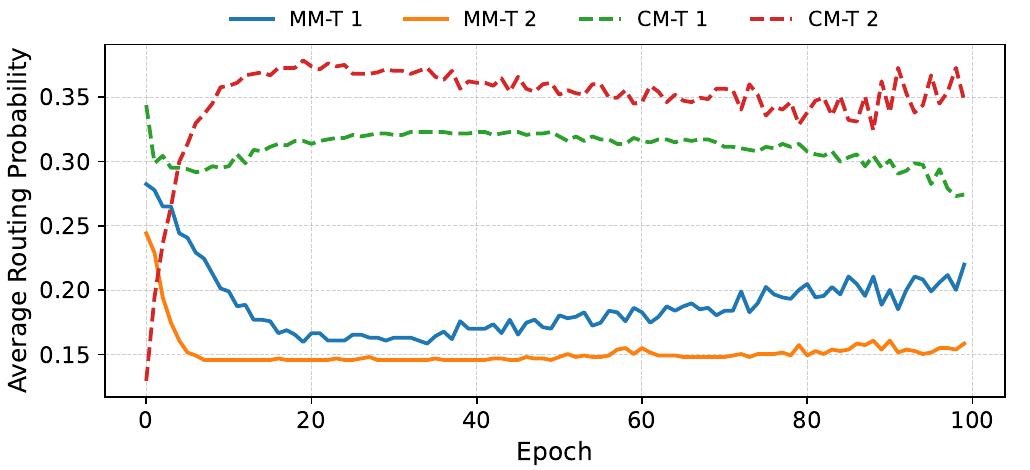}
	\caption{Average routing probabilities of specialized teachers from a single run of MST-Distill for the visual student on the RAVDESS dataset. Solid and dashed lines indicate multimodal and cross-modal teachers, respectively.}
	\label{fig:dynamic_mst}
\end{figure}

\subsubsection{Routing Dynamics in MST-Distill}
To better understand the mechanism of MST under dynamic knowledge, we further tracked the average routing probabilities of each teacher during a RAVDESS training run. As shown as Figure~\ref{fig:dynamic_mst}, both multimodal and cross-modal teachers demonstrated significant engagement with distinctive adaptive selection patterns throughout the training process. Notably, CM-T 2 initially showed minimal selection probability but gradually increased its contribution over time, clearly demonstrating the adaptive nature of our DKD strategy. This dynamic adjustment of teacher contributions validates the effectiveness of our approach in automatically identifying and leveraging the most valuable knowledge sources as training progresses.

\subsubsection{Configurations on MST-Distill}
To extend our analysis beyond basic component ablations, we conduct a detailed hyperparameter study of the MST-Distill framework, focusing on the nuanced configuration aspects within the Mixture of Specialized Teachers module. We systematically vary teacher compositions (cross-modal, multimodal, and their combinations) and examine how different values of the top-$k$ parameter affect the dynamic knowledge distillation process. These fine-grained experiments complement our main ablations by revealing the sensitivity of distillation effectiveness to specific parameter choices, providing practical insights for optimal deployment. Additional experimental results and detailed analyses can be found in Appendix~\ref{appendix:mst_others}.

\begin{figure}[htbp]
	\centering
	\includegraphics[width=\linewidth]{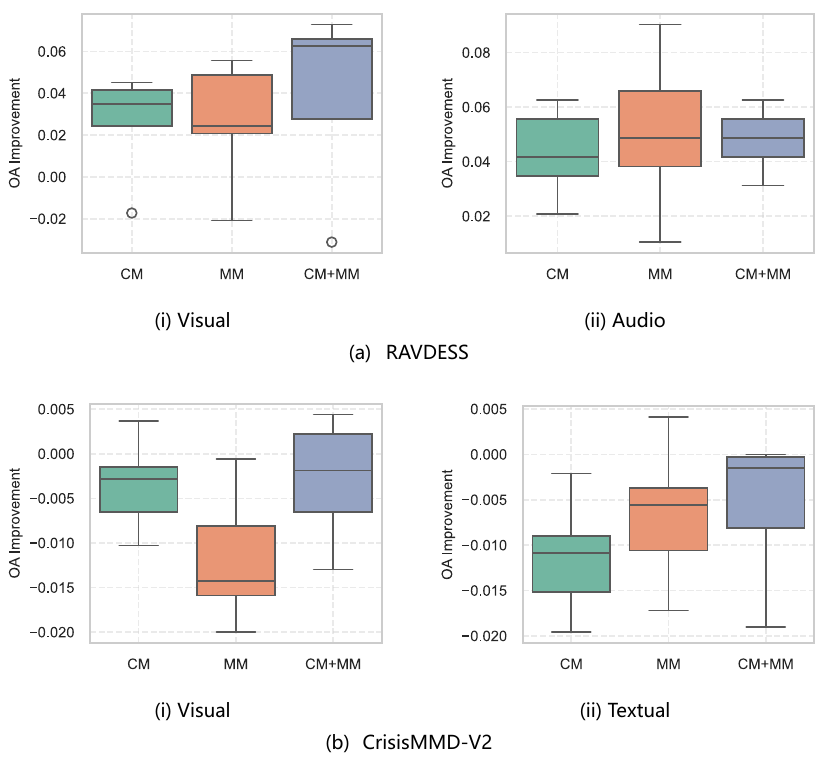}
	\caption{Box plots of OA improvements under different teacher configurations in MST. Results are based on five independent runs conducted on two representative multimodal classification datasets. Different box colors represent different teacher configurations.}
	\label{fig:mst_cfg}
\end{figure}

\paragraph{Effect of Teacher Diversity Configurations}
Experiments on two representative multimodal classification datasets (visual, audio, and textual modalities) demonstrate our approach's effectiveness. As Figure~\ref{fig:mst_cfg} shows, the combined teacher configuration (CM+MM) consistently outperforms individual cross-modal (CM) or multimodal (MM) settings, achieving higher median OA improvements with reduced variance. In RAVDESS, CM+MM delivers stable and significant gains, particularly for visual students, with compact interquartile ranges and minimal outliers. For CrisisMMD-V2, which features weakly aligned modalities, CM+MM maintains superior stability and effectiveness, especially for textual students. These findings confirm the robustness and generalizability of combining cross-modal and multimodal teacher guidance.

\begin{figure}[htbp]
	\centering
	\includegraphics[width=\linewidth]{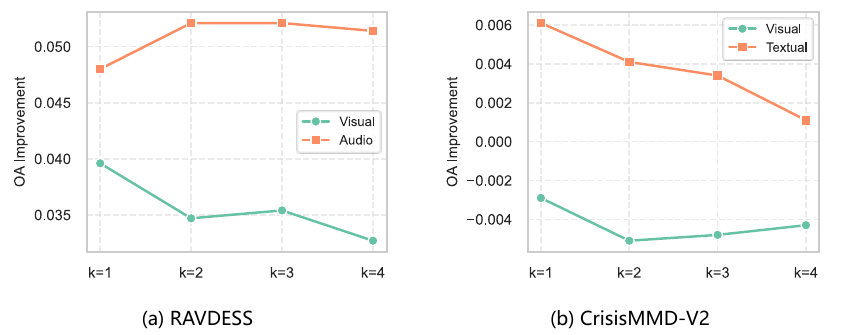}
	\caption{Performance trends under different top-$k$ values in dynamic knowledge distillation. Student performance across modalities is distinguished by color and marker.}
	\label{fig:dkd_k}
\end{figure}

\paragraph{Effect of Top-$k$ Teacher Selection}
Our sensitivity analysis examines parameter $k$, which controls the number of modality-specific teachers selected per sample during dynamic knowledge distillation. With $k$ ranging from 1 to 4 (limited by available feature layers for MaskNet insertion), Figure~\ref{fig:dkd_k} reveals that performance improves when $k$ is below the maximum value, while $k=4$ consistently underperforms. This occurs because the adaptive selection mechanism degrades into uniform averaging when all teachers are used, eliminating sample-specific discrimination. These findings validate our top-$k$-style solution. We adopt $k=1$ as our default across all experiments, as it consistently delivers robust performance on all datasets without requiring dataset-specific tuning.

\section{Conclusion and Discussion}
This paper proposes MST-Distill, a novel framework for cross-modal knowledge distillation that effectively tackles two critical challenges: distillation path selection and knowledge drift. Our approach incorporates a diverse ensemble of specialized teachers from both multimodal and cross-modal domains, utilizing an instance-level routing network to dynamically select optimal teachers for each input sample and a plug-in MaskNet module to address knowledge drift through response consistency supervision. Comprehensive evaluations across five diverse multimodal benchmarks demonstrate MST-Distill's superior performance and generalizability, consistently outperforming state-of-the-art knowledge distillation and mutual learning methods. While enhancing effectiveness for loosely aligned modalities remains an open challenge, this work establishes a foundation for leveraging teacher diversity in cross-modal knowledge transfer, with future research exploring more sophisticated methods including knowledge disentanglement and gradient modulation techniques, as well as extending to scenarios with three or more modalities.

\begin{acks}
This work was supported in part by the Shaanxi Key Technology R\&D Program under Grant 2024GX-ZDCYL-02-15 and the Natural Science Funds for Distinguished Young Scholar of Shaanxi under Grant 2025JC-JCQN-079.
\end{acks}

\clearpage
\bibliographystyle{ACM-Reference-Format}
\bibliography{MST-Distill}


\begin{thebibliography}{53}


\ifx \showCODEN    \undefined \def \showCODEN     #1{\unskip}     \fi
\ifx \showDOI      \undefined \def \showDOI       #1{#1}\fi
\ifx \showISBNx    \undefined \def \showISBNx     #1{\unskip}     \fi
\ifx \showISBNxiii \undefined \def \showISBNxiii  #1{\unskip}     \fi
\ifx \showISSN     \undefined \def \showISSN      #1{\unskip}     \fi
\ifx \showLCCN     \undefined \def \showLCCN      #1{\unskip}     \fi
\ifx \shownote     \undefined \def \shownote      #1{#1}          \fi
\ifx \showarticletitle \undefined \def \showarticletitle #1{#1}   \fi
\ifx \showURL      \undefined \def \showURL       {\relax}        \fi
\providecommand\bibfield[2]{#2}
\providecommand\bibinfo[2]{#2}
\providecommand\natexlab[1]{#1}
\providecommand\showeprint[2][]{arXiv:#2}

\bibitem[Alam et~al\mbox{.}(2018)]%
        {cmmdv2}
\bibfield{author}{\bibinfo{person}{Firoj Alam}, \bibinfo{person}{Ferda Ofli},
  {and} \bibinfo{person}{Muhammad Imran}.} \bibinfo{year}{2018}\natexlab{}.
\newblock \showarticletitle{CrisisMMD: Multimodal Twitter Datasets from Natural
  Disasters}. In \bibinfo{booktitle}{\emph{Proceedings of the International
  AAAI Conference on Web and Social Media}}, Vol.~\bibinfo{volume}{12}.
\newblock


\bibitem[Baltru{\v{s}}aitis et~al\mbox{.}(2018)]%
        {ref1}
\bibfield{author}{\bibinfo{person}{Tadas Baltru{\v{s}}aitis},
  \bibinfo{person}{Chaitanya Ahuja}, {and} \bibinfo{person}{Louis-Philippe
  Morency}.} \bibinfo{year}{2018}\natexlab{}.
\newblock \showarticletitle{Multimodal Machine Learning: A Survey and
  Taxonomy}.
\newblock \bibinfo{journal}{\emph{IEEE Transactions on Pattern Analysis and
  Machine Intelligence}} \bibinfo{volume}{41}, \bibinfo{number}{2}
  (\bibinfo{year}{2018}), \bibinfo{pages}{423--443}.
\newblock


\bibitem[Castrejon et~al\mbox{.}(2016)]%
        {ref9}
\bibfield{author}{\bibinfo{person}{Lluis Castrejon}, \bibinfo{person}{Yusuf
  Aytar}, \bibinfo{person}{Carl Vondrick}, \bibinfo{person}{Hamed Pirsiavash},
  {and} \bibinfo{person}{Antonio Torralba}.} \bibinfo{year}{2016}\natexlab{}.
\newblock \showarticletitle{Learning Aligned Cross-Modal Representations from
  Weakly Aligned Data}. In \bibinfo{booktitle}{\emph{2016 IEEE/CVF Conference
  on Computer Vision and Pattern Recognition (CVPR)}}.
  \bibinfo{pages}{2940--2949}.
\newblock


\bibitem[Chen et~al\mbox{.}(2020)]%
        {vggs}
\bibfield{author}{\bibinfo{person}{Honglie Chen}, \bibinfo{person}{Wjournali
  Xie}, \bibinfo{person}{Andrea Vedaldi}, {and} \bibinfo{person}{Andrew
  Zisserman}.} \bibinfo{year}{2020}\natexlab{}.
\newblock \showarticletitle{VGGSound: A Large-Scale Audio-Visual Dataset}. In
  \bibinfo{booktitle}{\emph{IEEE International Conference on Acoustics, Speech
  and Signal Processing (ICASSP)}}. IEEE, \bibinfo{pages}{721--725}.
\newblock


\bibitem[Choi and Lee(2019)]%
        {ref7}
\bibfield{author}{\bibinfo{person}{Jun-Ho Choi} {and}
  \bibinfo{person}{Jong-Seok Lee}.} \bibinfo{year}{2019}\natexlab{}.
\newblock \showarticletitle{EmbraceNet: A Robust Deep Learning Architecture for
  Multimodal Classification}.
\newblock \bibinfo{journal}{\emph{Information Fusion}}  \bibinfo{volume}{51}
  (\bibinfo{year}{2019}), \bibinfo{pages}{259--270}.
\newblock


\bibitem[Chung et~al\mbox{.}(2020)]%
        {ref15}
\bibfield{author}{\bibinfo{person}{Inseop Chung}, \bibinfo{person}{SeongUk
  Park}, \bibinfo{person}{Jangho Kim}, {and} \bibinfo{person}{Nojun Kwak}.}
  \bibinfo{year}{2020}\natexlab{}.
\newblock \showarticletitle{Feature-Map-Level Online Adversarial Knowledge
  Distillation}. In \bibinfo{booktitle}{\emph{International Conference on
  Machine Learning}}. \bibinfo{pages}{2006--2015}.
\newblock


\bibitem[Ding et~al\mbox{.}(2023)]%
        {ref34}
\bibfield{author}{\bibinfo{person}{Xinyi Ding}, \bibinfo{person}{Tao Han},
  \bibinfo{person}{Yili Fang}, {and} \bibinfo{person}{Eric Larson}.}
  \bibinfo{year}{2023}\natexlab{}.
\newblock \showarticletitle{An Approach for Combining Multimodal Fusion and
  Neural Architecture Search Applied to Knowledge Tracing}.
\newblock \bibinfo{journal}{\emph{Applied Intelligence}} \bibinfo{volume}{53},
  \bibinfo{number}{9} (\bibinfo{year}{2023}), \bibinfo{pages}{11092--11103}.
\newblock


\bibitem[Dong et~al\mbox{.}(2023)]%
        {ref37}
\bibfield{author}{\bibinfo{person}{Hao Dong}, \bibinfo{person}{Ismail Nejjar},
  \bibinfo{person}{Han Sun}, \bibinfo{person}{Eleni Chatzi}, {and}
  \bibinfo{person}{Olga Fink}.} \bibinfo{year}{2023}\natexlab{}.
\newblock \showarticletitle{SimMMDG: A Simple and Effective Framework for
  Multi-Modal Domain Generalization}.
\newblock \bibinfo{journal}{\emph{Advances in Neural Information Processing
  Systems}}  \bibinfo{volume}{36} (\bibinfo{year}{2023}),
  \bibinfo{pages}{78674--78695}.
\newblock


\bibitem[Fan et~al\mbox{.}(2023)]%
        {ref43}
\bibfield{author}{\bibinfo{person}{Yunfeng Fan}, \bibinfo{person}{Wenchao Xu},
  \bibinfo{person}{Haozhao Wang}, \bibinfo{person}{Junxiao Wang}, {and}
  \bibinfo{person}{Song Guo}.} \bibinfo{year}{2023}\natexlab{}.
\newblock \showarticletitle{PMR: Prototypical Modal Rebalance for Multimodal
  Learning}. In \bibinfo{booktitle}{\emph{2023 IEEE/CVF Conference on Computer
  Vision and Pattern Recognition (CVPR)}}. \bibinfo{pages}{20029--20038}.
\newblock


\bibitem[Ghamandi et~al\mbox{.}(2024)]%
        {ref5}
\bibfield{author}{\bibinfo{person}{Ryan~Khushan Ghamandi},
  \bibinfo{person}{Ravi~Kiran Kattoju}, \bibinfo{person}{Yahya Hmaiti},
  \bibinfo{person}{Mykola Maslych}, \bibinfo{person}{Eugene~Matthew Taranta},
  \bibinfo{person}{Ryan~P McMahan}, {and} \bibinfo{person}{Joseph LaViola}.}
  \bibinfo{year}{2024}\natexlab{}.
\newblock \showarticletitle{Unlocking Understanding: An Investigation of
  Multimodal Communication in Virtual Reality Collaboration}. In
  \bibinfo{booktitle}{\emph{Proceedings of the 2024 CHI Conference on Human
  Factors in Computing Systems}}. \bibinfo{pages}{1--16}.
\newblock


\bibitem[Gu et~al\mbox{.}(2024)]%
        {ref20}
\bibfield{author}{\bibinfo{person}{Yuxian Gu}, \bibinfo{person}{Li Dong},
  \bibinfo{person}{Furu Wei}, {and} \bibinfo{person}{Minlie Huang}.}
  \bibinfo{year}{2024}\natexlab{}.
\newblock \showarticletitle{MiniLLM: Knowledge Distillation of Large Language
  Models}. In \bibinfo{booktitle}{\emph{Proceedings of International Conference
  on Learning Representations}}.
\newblock


\bibitem[Hao et~al\mbox{.}(2023)]%
        {ofa}
\bibfield{author}{\bibinfo{person}{Zhiwei Hao}, \bibinfo{person}{Jianyuan Guo},
  \bibinfo{person}{Kai Han}, \bibinfo{person}{Yehui Tang}, \bibinfo{person}{Han
  Hu}, \bibinfo{person}{Yunhe Wang}, {and} \bibinfo{person}{Chang Xu}.}
  \bibinfo{year}{2023}\natexlab{}.
\newblock \showarticletitle{One-for-All: Bridge the Gap between Heterogeneous
  Architectures in Knowledge Distillation}.
\newblock \bibinfo{journal}{\emph{Advances in Neural Information Processing
  Systems}}  \bibinfo{volume}{36} (\bibinfo{year}{2023}),
  \bibinfo{pages}{79570--79582}.
\newblock


\bibitem[Hazirbas et~al\mbox{.}(2016)]%
        {fusenet}
\bibfield{author}{\bibinfo{person}{Caner Hazirbas}, \bibinfo{person}{Lingni
  Ma}, \bibinfo{person}{Csaba Domokos}, {and} \bibinfo{person}{Daniel
  Cremers}.} \bibinfo{year}{2016}\natexlab{}.
\newblock \showarticletitle{FuseNet: Incorporating Depth into Semantic
  Segmentation via Fusion-Based CNN Architecture}. In
  \bibinfo{booktitle}{\emph{Asian Conference on Computer Vision}}. Springer,
  \bibinfo{pages}{213--228}.
\newblock


\bibitem[He et~al\mbox{.}(2023)]%
        {ref38}
\bibfield{author}{\bibinfo{person}{Xiao He}, \bibinfo{person}{Chang Tang},
  \bibinfo{person}{Xin Zou}, {and} \bibinfo{person}{Wei Zhang}.}
  \bibinfo{year}{2023}\natexlab{}.
\newblock \showarticletitle{Multispectral Object Detection via Cross-Modal
  Conflict-Aware Learning}. In \bibinfo{booktitle}{\emph{Proceedings of the
  31st ACM International Conference on Multimedia}}.
  \bibinfo{pages}{1465--1474}.
\newblock


\bibitem[Hinton et~al\mbox{.}(2015)]%
        {kd}
\bibfield{author}{\bibinfo{person}{Geoffrey Hinton}, \bibinfo{person}{Oriol
  Vinyals}, {and} \bibinfo{person}{Jeff Dean}.}
  \bibinfo{year}{2015}\natexlab{}.
\newblock \bibinfo{title}{Distilling the Knowledge in a Neural Network}.
\newblock
\newblock
\showeprint[arxiv]{1503.02531}


\bibitem[Hu et~al\mbox{.}(2022)]%
        {mdecopling}
\bibfield{author}{\bibinfo{person}{Weipeng Hu}, \bibinfo{person}{Bohong Liu},
  \bibinfo{person}{Haitang Zeng}, \bibinfo{person}{Yanke Hou}, {and}
  \bibinfo{person}{Haifeng Hu}.} \bibinfo{year}{2022}\natexlab{}.
\newblock \showarticletitle{Adversarial Decoupling and Modality-Invariant
  Representation Learning for Visible-Infrared Person Re-Identification}.
\newblock \bibinfo{journal}{\emph{IEEE Transactions on Circuits and Systems for
  Video Technology}} \bibinfo{volume}{32}, \bibinfo{number}{8}
  (\bibinfo{year}{2022}), \bibinfo{pages}{5095--5109}.
\newblock


\bibitem[Huo et~al\mbox{.}(2024)]%
        {c2kd}
\bibfield{author}{\bibinfo{person}{Fushuo Huo}, \bibinfo{person}{Wenchao Xu},
  \bibinfo{person}{Jingcai Guo}, \bibinfo{person}{Haozhao Wang}, {and}
  \bibinfo{person}{Song Guo}.} \bibinfo{year}{2024}\natexlab{}.
\newblock \showarticletitle{C$^2$KD: Bridging the Modality Gap for Cross-Modal
  Knowledge Distillation}. In \bibinfo{booktitle}{\emph{2024 IEEE/CVF
  Conference on Computer Vision and Pattern Recognition (CVPR)}}.
  \bibinfo{pages}{16006--16015}.
\newblock


\bibitem[Jin et~al\mbox{.}(2023)]%
        {mlld}
\bibfield{author}{\bibinfo{person}{Ying Jin}, \bibinfo{person}{Jiaqi Wang},
  {and} \bibinfo{person}{Dahua Lin}.} \bibinfo{year}{2023}\natexlab{}.
\newblock \showarticletitle{Multi-level logit distillation}. In
  \bibinfo{booktitle}{\emph{Proceedings of the IEEE/CVF Conference on Computer
  Vision and Pattern Recognition (CVPR)}}. \bibinfo{pages}{24276--24285}.
\newblock


\bibitem[Kim and Kang(2022)]%
        {cmkd_app1}
\bibfield{author}{\bibinfo{person}{Donghwa Kim} {and} \bibinfo{person}{Pilsung
  Kang}.} \bibinfo{year}{2022}\natexlab{}.
\newblock \showarticletitle{Cross-Modal Distillation with Audio-Text Fusion for
  Fine-Grained Emotion Classification Using BERT and Wav2vec 2.0}.
\newblock \bibinfo{journal}{\emph{Neurocomputing}}  \bibinfo{volume}{506}
  (\bibinfo{year}{2022}), \bibinfo{pages}{168--183}.
\newblock


\bibitem[Kleinman et~al\mbox{.}(2023)]%
        {ref40}
\bibfield{author}{\bibinfo{person}{Michael Kleinman},
  \bibinfo{person}{Alessandro Achille}, {and} \bibinfo{person}{Stefano
  Soatto}.} \bibinfo{year}{2023}\natexlab{}.
\newblock \showarticletitle{Critical Learning Periods for Multisensory
  Integration in Deep Networks}. In \bibinfo{booktitle}{\emph{2023 IEEE/CVF
  Conference on Computer Vision and Pattern Recognition (CVPR)}}.
  \bibinfo{pages}{24296--24305}.
\newblock


\bibitem[Li et~al\mbox{.}(2024a)]%
        {ref33}
\bibfield{author}{\bibinfo{person}{Ke Li}, \bibinfo{person}{Fuyu Dong},
  \bibinfo{person}{Di Wang}, \bibinfo{person}{Shaofeng Li},
  \bibinfo{person}{Quan Wang}, \bibinfo{person}{Xinbo Gao}, {and}
  \bibinfo{person}{Tat-Seng Chua}.} \bibinfo{year}{2024}\natexlab{a}.
\newblock \bibinfo{title}{Show Me What and Where Has Changed? Question
  Answering and Grounding for Remote Sensing Change Detection}.
\newblock
\newblock
\showeprint[arxiv]{2410.23828}


\bibitem[Li et~al\mbox{.}(2024b)]%
        {ref24}
\bibfield{author}{\bibinfo{person}{Ke Li}, \bibinfo{person}{Di Wang},
  \bibinfo{person}{Haojie Xu}, \bibinfo{person}{Haodi Zhong}, {and}
  \bibinfo{person}{Cong Wang}.} \bibinfo{year}{2024}\natexlab{b}.
\newblock \showarticletitle{Language-Guided Progressive Attention for Visual
  Grounding in Remote Sensing Images}.
\newblock \bibinfo{journal}{\emph{IEEE Transactions on Geoscience and Remote
  Sensing}} (\bibinfo{year}{2024}).
\newblock


\bibitem[Li et~al\mbox{.}(2023)]%
        {od}
\bibfield{author}{\bibinfo{person}{Zhihui Li}, \bibinfo{person}{Pengfei Xu},
  \bibinfo{person}{Xiaojun Chang}, \bibinfo{person}{Luyao Yang},
  \bibinfo{person}{Yuanyuan Zhang}, \bibinfo{person}{Lina Yao}, {and}
  \bibinfo{person}{Xiaojiang Chen}.} \bibinfo{year}{2023}\natexlab{}.
\newblock \showarticletitle{When Object Detection Meets Knowledge Distillation:
  A Survey}.
\newblock \bibinfo{journal}{\emph{IEEE Transactions on Pattern Analysis and
  Machine Intelligence}} \bibinfo{volume}{45}, \bibinfo{number}{8}
  (\bibinfo{year}{2023}), \bibinfo{pages}{10555--10579}.
\newblock


\bibitem[Liang et~al\mbox{.}(2024)]%
        {ref28}
\bibfield{author}{\bibinfo{person}{Xiao Liang}, \bibinfo{person}{Yanlei Zhang},
  \bibinfo{person}{Di Wang}, \bibinfo{person}{Haodi Zhong},
  \bibinfo{person}{Ronghan Li}, {and} \bibinfo{person}{Quan Wang}.}
  \bibinfo{year}{2024}\natexlab{}.
\newblock \showarticletitle{Divide and Conquer: Isolating Normal-Abnormal
  Attributes in Knowledge Graph-Enhanced Radiology Report Generation}. In
  \bibinfo{booktitle}{\emph{Proceedings of the 32nd ACM International
  Conference on Multimedia}}. \bibinfo{pages}{4967--4975}.
\newblock


\bibitem[Liang et~al\mbox{.}(2022)]%
        {nmmm}
\bibfield{author}{\bibinfo{person}{Yupeng Liang}, \bibinfo{person}{Ryosuke
  Wakaki}, \bibinfo{person}{Shohei Nobuhara}, {and} \bibinfo{person}{Ko
  Nishino}.} \bibinfo{year}{2022}\natexlab{}.
\newblock \showarticletitle{Multimodal Material Segmentation}. In
  \bibinfo{booktitle}{\emph{2022 IEEE/CVF Conference on Computer Vision and
  Pattern Recognition (CVPR)}}. \bibinfo{pages}{19800--19808}.
\newblock


\bibitem[Livingstone and Russo(2018)]%
        {ravdess}
\bibfield{author}{\bibinfo{person}{Steven~R Livingstone} {and}
  \bibinfo{person}{Frank~A Russo}.} \bibinfo{year}{2018}\natexlab{}.
\newblock \showarticletitle{The Ryerson Audio-Visual Database of Emotional
  Speech and Song (RAVDESS): A Dynamic, Multimodal Set of Facial and Vocal
  Expressions in North American English}.
\newblock \bibinfo{journal}{\emph{PloS one}} \bibinfo{volume}{13},
  \bibinfo{number}{5} (\bibinfo{year}{2018}), \bibinfo{pages}{e0196391}.
\newblock


\bibitem[Ma et~al\mbox{.}(2023)]%
        {ref6}
\bibfield{author}{\bibinfo{person}{Mengmeng Ma}, \bibinfo{person}{Jian Ren},
  \bibinfo{person}{Long Zhao}, \bibinfo{person}{Davide Testuggine}, {and}
  \bibinfo{person}{Xi Peng}.} \bibinfo{year}{2023}\natexlab{}.
\newblock \showarticletitle{Are Multimodal Transformers Robust to Missing
  Modality?}. In \bibinfo{booktitle}{\emph{2023 IEEE/CVF Conference on Computer
  Vision and Pattern Recognition (CVPR)}}. \bibinfo{pages}{18177--18186}.
\newblock


\bibitem[Ma et~al\mbox{.}(2024)]%
        {metalearning}
\bibfield{author}{\bibinfo{person}{Wenxuan Ma}, \bibinfo{person}{Shuang Li},
  \bibinfo{person}{Lincan Cai}, {and} \bibinfo{person}{Jingxuan Kang}.}
  \bibinfo{year}{2024}\natexlab{}.
\newblock \showarticletitle{Learning Modality Knowledge Alignment for
  Cross-Modality Transfer}. In \bibinfo{booktitle}{\emph{Proceedings of the
  41st International Conference on Machine Learning}}.
  \bibinfo{pages}{33777--33793}.
\newblock


\bibitem[Ngiam et~al\mbox{.}(2011)]%
        {ref8}
\bibfield{author}{\bibinfo{person}{Jiquan Ngiam}, \bibinfo{person}{Aditya
  Khosla}, \bibinfo{person}{Mingyu Kim}, \bibinfo{person}{Juhan Nam},
  \bibinfo{person}{Honglak Lee}, \bibinfo{person}{Andrew~Y Ng},
  {et~al\mbox{.}}} \bibinfo{year}{2011}\natexlab{}.
\newblock \showarticletitle{Multimodal Deep Learning}. In
  \bibinfo{booktitle}{\emph{International Conference on Machine Learning}},
  Vol.~\bibinfo{volume}{11}. \bibinfo{pages}{689--696}.
\newblock


\bibitem[Park et~al\mbox{.}(2019)]%
        {rkd}
\bibfield{author}{\bibinfo{person}{Wonpyo Park}, \bibinfo{person}{Dongju Kim},
  \bibinfo{person}{Yan Lu}, {and} \bibinfo{person}{Minsu Cho}.}
  \bibinfo{year}{2019}\natexlab{}.
\newblock \showarticletitle{Relational Knowledge Distillation}. In
  \bibinfo{booktitle}{\emph{2019 IEEE/CVF Conference on Computer Vision and
  Pattern Recognition (CVPR)}}. \bibinfo{pages}{3967--3976}.
\newblock


\bibitem[Ren et~al\mbox{.}(2024)]%
        {ref31}
\bibfield{author}{\bibinfo{person}{Shuhuai Ren}, \bibinfo{person}{Linli Yao},
  \bibinfo{person}{Shicheng Li}, \bibinfo{person}{Xu Sun}, {and}
  \bibinfo{person}{Lu Hou}.} \bibinfo{year}{2024}\natexlab{}.
\newblock \showarticletitle{TimeChat: A Time-Sensitive Multimodal Large
  Language Model for Long Video Understanding}. In
  \bibinfo{booktitle}{\emph{2024 IEEE/CVF Conference on Computer Vision and
  Pattern Recognition (CVPR)}}. \bibinfo{pages}{14313--14323}.
\newblock


\bibitem[Romero et~al\mbox{.}(2015)]%
        {fitnets}
\bibfield{author}{\bibinfo{person}{Adriana Romero}, \bibinfo{person}{Nicolas
  Ballas}, \bibinfo{person}{Samira~Ebrahimi Kahou}, \bibinfo{person}{Antoine
  Chassang}, \bibinfo{person}{Carlo Gatta}, {and} \bibinfo{person}{Yoshua
  Bengio}.} \bibinfo{year}{2015}\natexlab{}.
\newblock \showarticletitle{FitNets: Hints for Thin Deep Nets}. In
  \bibinfo{booktitle}{\emph{Proceedings of International Conference on Learning
  Representations}}.
\newblock


\bibitem[Selvaraju et~al\mbox{.}(2017)]%
        {grad_cam}
\bibfield{author}{\bibinfo{person}{Ramprasaath~R Selvaraju},
  \bibinfo{person}{Michael Cogswell}, \bibinfo{person}{Abhishek Das},
  \bibinfo{person}{Ramakrishna Vedantam}, \bibinfo{person}{Devi Parikh}, {and}
  \bibinfo{person}{Dhruv Batra}.} \bibinfo{year}{2017}\natexlab{}.
\newblock \showarticletitle{Grad-CAM: Visual Explanations from Deep Networks
  via Gradient-Based Localization}. In \bibinfo{booktitle}{\emph{Proceedings of
  the IEEE International Conference on Computer Vision}}.
  \bibinfo{pages}{618--626}.
\newblock


\bibitem[Silberman et~al\mbox{.}(2012)]%
        {nyudv2}
\bibfield{author}{\bibinfo{person}{Nathan Silberman}, \bibinfo{person}{Derek
  Hoiem}, \bibinfo{person}{Pushmeet Kohli}, {and} \bibinfo{person}{Rob
  Fergus}.} \bibinfo{year}{2012}\natexlab{}.
\newblock \showarticletitle{Indoor Segmentation and Support Inference from RGBD
  Images}. In \bibinfo{booktitle}{\emph{European Conference on Computer
  Vision}}. Springer, \bibinfo{pages}{746--760}.
\newblock


\bibitem[Stanton et~al\mbox{.}(2021)]%
        {kd_insights}
\bibfield{author}{\bibinfo{person}{Samuel Stanton}, \bibinfo{person}{Pavel
  Izmailov}, \bibinfo{person}{Polina Kirichenko}, \bibinfo{person}{Alexander~A
  Alemi}, {and} \bibinfo{person}{Andrew~G Wilson}.}
  \bibinfo{year}{2021}\natexlab{}.
\newblock \showarticletitle{Does Knowledge Distillation Really Work?}. In
  \bibinfo{booktitle}{\emph{Advances in Neural Information Processing
  Systems}}, Vol.~\bibinfo{volume}{34}. \bibinfo{pages}{6906--6919}.
\newblock


\bibitem[Sun et~al\mbox{.}(2023)]%
        {ref30}
\bibfield{author}{\bibinfo{person}{Jun Sun}, \bibinfo{person}{Shoukang Han},
  \bibinfo{person}{Yu-Ping Ruan}, \bibinfo{person}{Xiaoning Zhang},
  \bibinfo{person}{Shu-Kai Zheng}, \bibinfo{person}{Yulong Liu},
  \bibinfo{person}{Yuxin Huang}, {and} \bibinfo{person}{Taihao Li}.}
  \bibinfo{year}{2023}\natexlab{}.
\newblock \showarticletitle{Layer-Wise Fusion with Modality Independence
  Modeling for Multi-Modal Emotion Recognition}. In
  \bibinfo{booktitle}{\emph{Annual Meeting of the Association for Computational
  Linguistics}}. \bibinfo{pages}{658--670}.
\newblock


\bibitem[Tian et~al\mbox{.}(2020)]%
        {crd}
\bibfield{author}{\bibinfo{person}{Yonglong Tian}, \bibinfo{person}{Dilip
  Krishnan}, {and} \bibinfo{person}{Phillip Isola}.}
  \bibinfo{year}{2020}\natexlab{}.
\newblock \showarticletitle{Contrastive Representation Distillation}. In
  \bibinfo{booktitle}{\emph{Proceedings of International Conference on Learning
  Representations}}.
\newblock


\bibitem[Vaswani et~al\mbox{.}(2017)]%
        {transformer}
\bibfield{author}{\bibinfo{person}{Ashish Vaswani}, \bibinfo{person}{Noam
  Shazeer}, \bibinfo{person}{Niki Parmar}, \bibinfo{person}{Jakob Uszkoreit},
  \bibinfo{person}{Llion Jones}, \bibinfo{person}{Aidan~N Gomez},
  \bibinfo{person}{{\L}ukasz Kaiser}, {and} \bibinfo{person}{Illia
  Polosukhin}.} \bibinfo{year}{2017}\natexlab{}.
\newblock \showarticletitle{Attention Is All You Need}.
\newblock \bibinfo{journal}{\emph{Advances in Neural Information Processing
  Systems}}  \bibinfo{volume}{30} (\bibinfo{year}{2017}).
\newblock


\bibitem[Vielzeuf et~al\mbox{.}(2018)]%
        {avmnist}
\bibfield{author}{\bibinfo{person}{Valentin Vielzeuf}, \bibinfo{person}{Alexis
  Lechervy}, \bibinfo{person}{St{\'e}phane Pateux}, {and}
  \bibinfo{person}{Fr{\'e}d{\'e}ric Jurie}.} \bibinfo{year}{2018}\natexlab{}.
\newblock \showarticletitle{CentralNet: A Multilayer Approach for Multimodal
  Fusion}. In \bibinfo{booktitle}{\emph{European Conference on Computer Vision
  Workshops}}.
\newblock


\bibitem[Wang et~al\mbox{.}(2022)]%
        {ref4}
\bibfield{author}{\bibinfo{person}{Di Wang}, \bibinfo{person}{Caiping Zhang},
  \bibinfo{person}{Quan Wang}, \bibinfo{person}{Yumin Tian},
  \bibinfo{person}{Lihuo He}, {and} \bibinfo{person}{Lin Zhao}.}
  \bibinfo{year}{2022}\natexlab{}.
\newblock \showarticletitle{Hierarchical Semantic Structure Preserving Hashing
  for Cross-Modal Retrieval}.
\newblock \bibinfo{journal}{\emph{IEEE Transactions on Multimedia}}
  \bibinfo{volume}{25} (\bibinfo{year}{2022}), \bibinfo{pages}{1217--1229}.
\newblock


\bibitem[Wang et~al\mbox{.}(2024)]%
        {ref3}
\bibfield{author}{\bibinfo{person}{Yi Wang}, \bibinfo{person}{Kunchang Li},
  \bibinfo{person}{Xinhao Li}, \bibinfo{person}{Jiashuo Yu},
  \bibinfo{person}{Yinan He}, \bibinfo{person}{Guo Chen},
  \bibinfo{person}{Baoqi Pei}, \bibinfo{person}{Rongkun Zheng},
  \bibinfo{person}{Zun Wang}, \bibinfo{person}{Yansong Shi}, {et~al\mbox{.}}}
  \bibinfo{year}{2024}\natexlab{}.
\newblock \showarticletitle{InternVideo2: Scaling Foundation Models for
  Multimodal Video Understanding}. In \bibinfo{booktitle}{\emph{European
  Conference on Computer Vision}}. Springer, \bibinfo{pages}{396--416}.
\newblock


\bibitem[Wei and Hu(2024)]%
        {mmpareto}
\bibfield{author}{\bibinfo{person}{Yake Wei} {and} \bibinfo{person}{Di Hu}.}
  \bibinfo{year}{2024}\natexlab{}.
\newblock \showarticletitle{MMPareto: Boosting Multimodal Learning with
  Innocent Unimodal Assistance}. In \bibinfo{booktitle}{\emph{International
  Conference on Machine Learning}}. PMLR, \bibinfo{pages}{52559--52572}.
\newblock


\bibitem[Wu et~al\mbox{.}(2022)]%
        {ref39}
\bibfield{author}{\bibinfo{person}{Nan Wu}, \bibinfo{person}{Stanislaw
  Jastrzebski}, \bibinfo{person}{Kyunghyun Cho}, {and}
  \bibinfo{person}{Krzysztof~J Geras}.} \bibinfo{year}{2022}\natexlab{}.
\newblock \showarticletitle{Characterizing and Overcoming the Greedy Nature of
  Learning in Multi-Modal Deep Neural Networks}. In
  \bibinfo{booktitle}{\emph{International Conference on Machine Learning}}.
  PMLR, \bibinfo{pages}{24043--24055}.
\newblock


\bibitem[Xue et~al\mbox{.}(2023)]%
        {mfh}
\bibfield{author}{\bibinfo{person}{Zihui Xue}, \bibinfo{person}{Zhengqi Gao},
  \bibinfo{person}{Sucheng Ren}, {and} \bibinfo{person}{Hang Zhao}.}
  \bibinfo{year}{2023}\natexlab{}.
\newblock \showarticletitle{The Modality Focusing Hypothesis: Towards
  Understanding Crossmodal Knowledge Distillation}. In
  \bibinfo{booktitle}{\emph{Proceedings of International Conference on Learning
  Representations}}.
\newblock


\bibitem[Xue et~al\mbox{.}(2021)]%
        {mke}
\bibfield{author}{\bibinfo{person}{Zihui Xue}, \bibinfo{person}{Sucheng Ren},
  \bibinfo{person}{Zhengqi Gao}, {and} \bibinfo{person}{Hang Zhao}.}
  \bibinfo{year}{2021}\natexlab{}.
\newblock \showarticletitle{Multimodal Knowledge Expansion}. In
  \bibinfo{booktitle}{\emph{2021 IEEE/CVF Conference on Computer Vision and
  Pattern Recognition (CVPR)}}. \bibinfo{pages}{854--863}.
\newblock


\bibitem[Yang et~al\mbox{.}(2023)]%
        {ref27}
\bibfield{author}{\bibinfo{person}{Antoine Yang}, \bibinfo{person}{Arsha
  Nagrani}, \bibinfo{person}{Paul~Hongsuck Seo}, \bibinfo{person}{Antoine
  Miech}, \bibinfo{person}{Jordi Pont-Tuset}, \bibinfo{person}{Ivan Laptev},
  \bibinfo{person}{Josef Sivic}, {and} \bibinfo{person}{Cordelia Schmid}.}
  \bibinfo{year}{2023}\natexlab{}.
\newblock \showarticletitle{Vid2Seq: Large-Scale Pretraining of a Visual
  Language Model for Dense Video Captioning}. In \bibinfo{booktitle}{\emph{2023
  IEEE/CVF Conference on Computer Vision and Pattern Recognition (CVPR)}}.
  \bibinfo{pages}{10714--10726}.
\newblock


\bibitem[Yao et~al\mbox{.}(2022)]%
        {ecc}
\bibfield{author}{\bibinfo{person}{Jiangchao Yao}, \bibinfo{person}{Shengyu
  Zhang}, \bibinfo{person}{Yang Yao}, \bibinfo{person}{Feng Wang},
  \bibinfo{person}{Jianxin Ma}, \bibinfo{person}{Jianwei Zhang},
  \bibinfo{person}{Yunfei Chu}, \bibinfo{person}{Luo Ji},
  \bibinfo{person}{Kunyang Jia}, \bibinfo{person}{Tao Shen}, {et~al\mbox{.}}}
  \bibinfo{year}{2022}\natexlab{}.
\newblock \showarticletitle{Edge-Cloud Polarization and Collaboration: A
  Comprehensive Survey for AI}.
\newblock \bibinfo{journal}{\emph{IEEE Transactions on Knowledge and Data
  Engineering}} \bibinfo{volume}{35}, \bibinfo{number}{7}
  (\bibinfo{year}{2022}), \bibinfo{pages}{6866--6886}.
\newblock


\bibitem[Yim et~al\mbox{.}(2017)]%
        {fsp}
\bibfield{author}{\bibinfo{person}{Junho Yim}, \bibinfo{person}{Donggyu Joo},
  \bibinfo{person}{Jihoon Bae}, {and} \bibinfo{person}{Junmo Kim}.}
  \bibinfo{year}{2017}\natexlab{}.
\newblock \showarticletitle{A Gift from Knowledge Distillation: Fast
  Optimization, Network Minimization and Transfer Learning}. In
  \bibinfo{booktitle}{\emph{2017 IEEE/CVF Conference on Computer Vision and
  Pattern Recognition (CVPR)}}. \bibinfo{pages}{4133--4141}.
\newblock


\bibitem[Zhang and Wu(2022)]%
        {cmkd_ssr}
\bibfield{author}{\bibinfo{person}{Li Zhang} {and} \bibinfo{person}{Xiangqian
  Wu}.} \bibinfo{year}{2022}\natexlab{}.
\newblock \showarticletitle{Latent Space Semantic Supervision Based on
  Knowledge Distillation for Cross-Modal Retrieval}.
\newblock \bibinfo{journal}{\emph{IEEE Transactions on Image Processing}}
  \bibinfo{volume}{31} (\bibinfo{year}{2022}), \bibinfo{pages}{7154--7164}.
\newblock


\bibitem[Zhang et~al\mbox{.}(2024)]%
        {unim_bias}
\bibfield{author}{\bibinfo{person}{Yedi Zhang}, \bibinfo{person}{Peter Latham},
  {et~al\mbox{.}}} \bibinfo{year}{2024}\natexlab{}.
\newblock \showarticletitle{Understanding Unimodal Bias in Multimodal Deep
  Linear Networks}. In \bibinfo{booktitle}{\emph{International Conference on
  Machine Learning}}, Vol.~\bibinfo{volume}{235}. PMLR.
\newblock


\bibitem[Zhang et~al\mbox{.}(2018)]%
        {dml}
\bibfield{author}{\bibinfo{person}{Ying Zhang}, \bibinfo{person}{Tao Xiang},
  \bibinfo{person}{Timothy~M Hospedales}, {and} \bibinfo{person}{Huchuan Lu}.}
  \bibinfo{year}{2018}\natexlab{}.
\newblock \showarticletitle{Deep Mutual Learning}. In
  \bibinfo{booktitle}{\emph{2018 IEEE/CVF Conference on Computer Vision and
  Pattern Recognition (CVPR)}}. \bibinfo{pages}{4320--4328}.
\newblock


\bibitem[Zhao et~al\mbox{.}(2017)]%
        {ref2}
\bibfield{author}{\bibinfo{person}{Jing Zhao}, \bibinfo{person}{Xijiong Xie},
  \bibinfo{person}{Xin Xu}, {and} \bibinfo{person}{Shiliang Sun}.}
  \bibinfo{year}{2017}\natexlab{}.
\newblock \showarticletitle{Multi-View Learning Overview: Recent Progress and
  New Challenges}.
\newblock \bibinfo{journal}{\emph{Information Fusion}}  \bibinfo{volume}{38}
  (\bibinfo{year}{2017}), \bibinfo{pages}{43--54}.
\newblock


\bibitem[Zhou et~al\mbox{.}(2022)]%
        {vggs-50k}
\bibfield{author}{\bibinfo{person}{Jinxing Zhou}, \bibinfo{person}{Dan Guo},
  {and} \bibinfo{person}{Meng Wang}.} \bibinfo{year}{2022}\natexlab{}.
\newblock \showarticletitle{Contrastive Positive Sample Propagation along the
  Audio-Visual Event Line}.
\newblock \bibinfo{journal}{\emph{IEEE Transactions on Pattern Analysis and
  Machine Intelligence}} \bibinfo{volume}{45}, \bibinfo{number}{6}
  (\bibinfo{year}{2022}), \bibinfo{pages}{7239--7257}.
\newblock


\end{thebibliography}

\clearpage
\appendix

\section{Supplementary Analysis Supporting Method Motivation}
\label{appendix:moti}

\subsection{Distillation Path Selection}
As discussed in the main text, cross-modal knowledge distillation suffers from inherent asymmetry and uncertainty, which manifest in two aspects: the knowledge transferability between unimodal models, and the transferability from multimodal to unimodal models. Further experiments and analysis are provided in the following sections.

\begin{figure}[htbp]
	\centering
	\includegraphics[width=\linewidth]{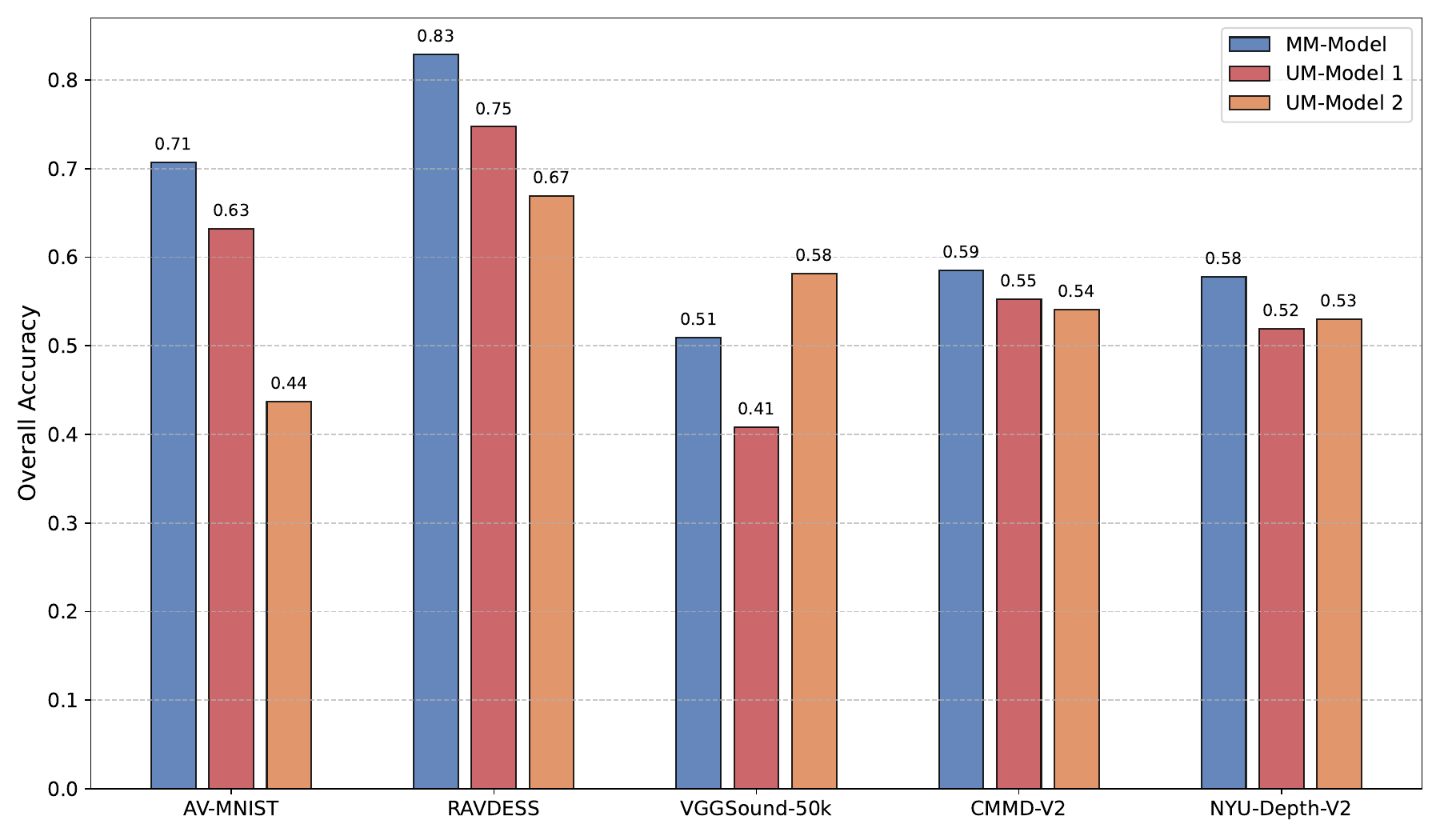}
	\caption{Performance comparison of multimodal and unimodal models trained from scratch across five multimodal datasets. Different colors are used to distinguish the performance of the multimodal model and various unimodal models.}
	\label{fig:appendix_ts}
\end{figure}

\subsubsection{Potential advantages of multimodal teachers.}
We independently train the multimodal and unimodal models corresponding to the five datasets mentioned in the main text, in order to establish performance benchmarks for the teacher and student models. The results are visualized as grouped bar charts in Figure~\ref{fig:appendix_ts}. It can be clearly observed that, in most cases, multimodal models effectively integrate information from different modalities and achieve better performance than unimodal models—except on the VGGSound-50k dataset, where a significant modality imbalance exists. This suggests that multimodal models are capable of learning complementary prior knowledge across modalities, which gives them a potential advantage in cross-modal knowledge distillation tasks. This also motivates our utilization of multimodal teacher models within the conventional cross-modal knowledge distillation framework.

\begin{figure}[htbp]
	\centering
	\includegraphics[width=\linewidth]{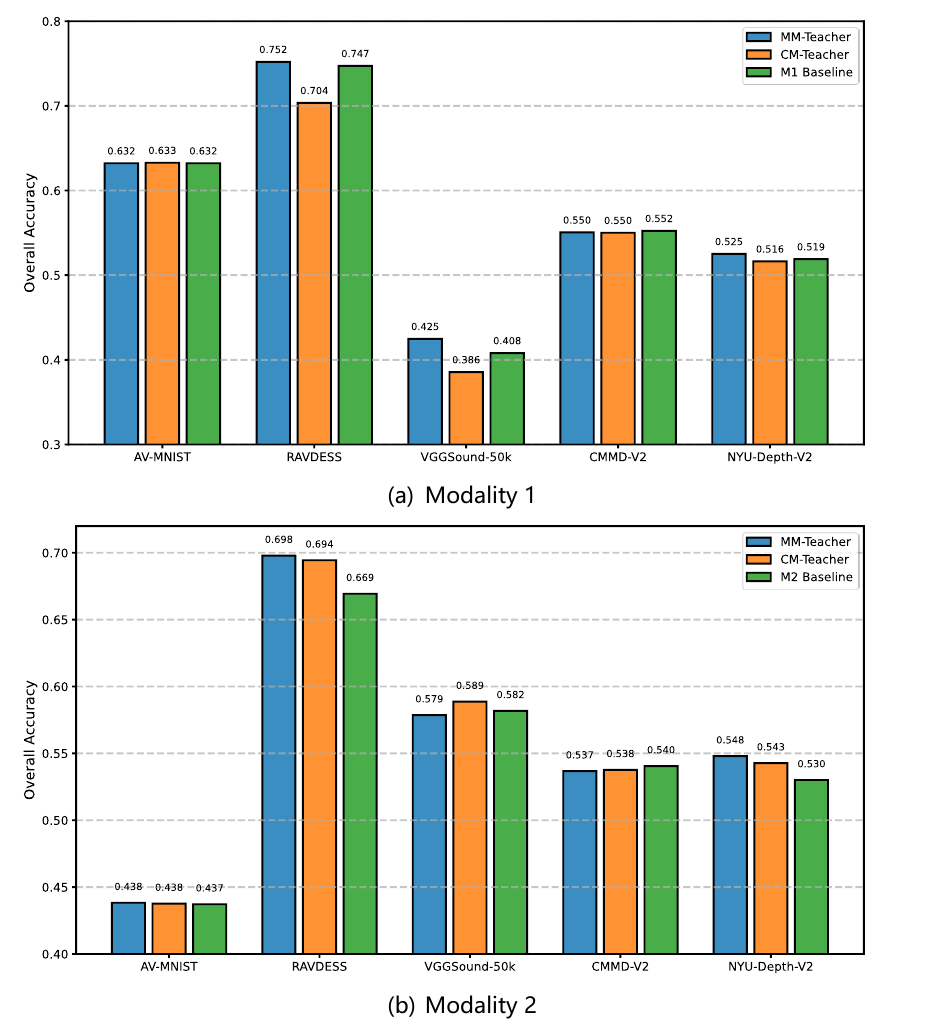}
	\caption{Performance comparison between multimodal and cross-modal teachers under response-based knowledge distillation. Subfigures (a) and (b) present the global accuracies of target-modality student models under different teacher selection strategies. The performance of unimodal models trained from scratch is indicated in green for reference.}
	\label{fig:appendix_kd}
\end{figure}

\subsubsection{Knowledge transferability differences among diverse teachers.}
We further conduct teacher diversity experiments using the classic logits-based knowledge distillation method across the same datasets. The overall accuracies of different teacher-student combinations are shown as bar charts in Figure~\ref{fig:appendix_kd}.

Multimodal teacher models demonstrate strong transferability across all modalities on the RAVDESS and NYU-Depth-V2 datasets, and exhibit good transferability specifically to the visual modality on VGGSound-50k. However, their effectiveness is not significant on the remaining datasets. In summary, while multimodal teachers show potential for cross-modal knowledge distillation, their success is not guaranteed—reflecting the uncertainty previously discussed.

This variability in transferability also exists among cross-modal teachers, and often manifests as asymmetry. For instance, on the RAVDESS and VGGSound datasets, knowledge distillation from audio teachers to visual students fails, whereas the reverse direction proves to be effective. It is worth noting that such disadvantages are difficult to overcome in conventional cross-modal distillation settings that rely solely on a single cross-modal teacher.

These observations highlight that relying on either a single multimodal teacher or a single cross-modal teacher alone is insufficient for consistently achieving optimal distillation outcomes, thereby giving rise to the distillation path selection problem in cross-modal knowledge distillation.

\begin{figure*}[htbp]
	\centering
	\includegraphics[width=\linewidth]{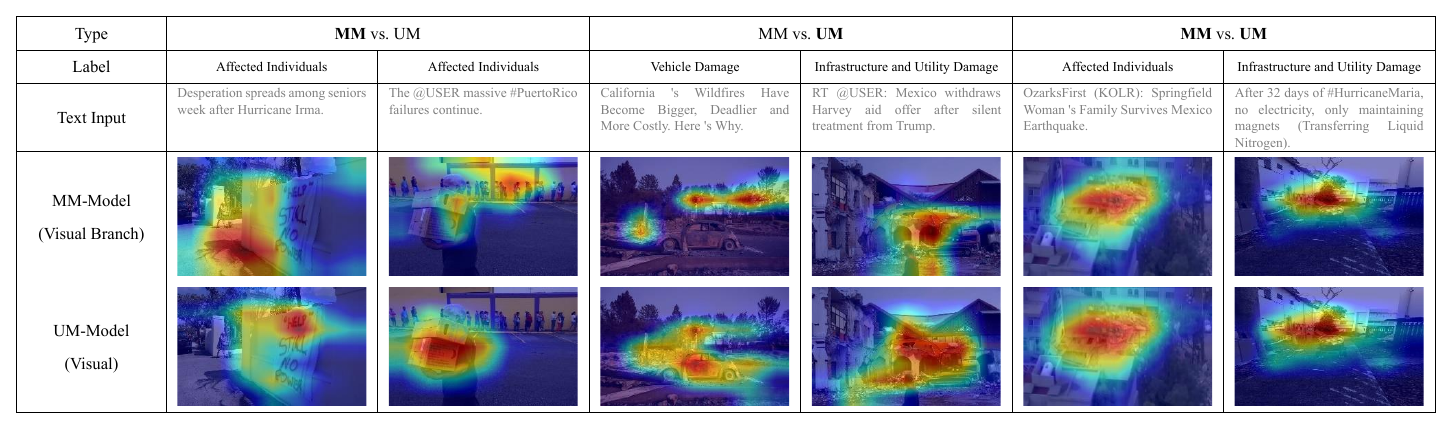}
	\caption{Grad-CAM visualizations comparing the visual attention of a multimodal teacher and a unimodal visual student on the CrisisMMD-V2 dataset. Text inputs are shown above each image. The attention map of the better-performing model is highlighted in bold; both are bolded if their performance is similar.}
	\label{fig:appendix_drift}
\end{figure*}

\subsection{Knowledge Drift}
As mentioned in the main text, the difference in inductive biases under different modality inputs not only arises between distinct unimodal models, but also exists between the target-modality branch of a multimodal model and its corresponding unimodal model. To illustrate this, we present Grad-CAM visualizations (Figure~\ref{fig:appendix_drift}) comparing the visual branch of a multimodal teacher model and a visual student model, both trained from scratch on the CrisisMMD-V2 dataset.

The behavioral differences observed between these models exhibit notable complexity and task-dependent variations. From a performance perspective, the visual branch of the multimodal model demonstrates superior performance in certain scenarios. Specifically, in the first two columns, the multimodal visual branch accurately focuses on the target person within the image, whereas the unimodal visual student fails to achieve this precise localization. Conversely, the unimodal student exhibits superior performance in alternative contexts. As evidenced in the middle two columns, the multimodal model erroneously attends to irrelevant regions including background elements and non-target individuals, while the unimodal model maintains correct focus on the intended subject. In some instances, the performance differences between the two approaches are marginal, suggesting task-specific advantages rather than consistent superiority of either method.

These findings indicate that such inconsistencies in model behavior occur at the instance level, which can result in knowledge drift between the teacher and student models. This further motivates the introduction of MaskNet, an instance-aware and learnable teacher feature reconstruction module designed to adaptively align the teacher's guidance with the needs of each student instance.

\section{Supplementary Implementation and Experimental Details}

\subsection{Dataset Descriptions}
\label{appendix:dataset}
For the multimodal datasets used in the main text, we provide a summary of their key characteristics in Table~\ref{tab:appendix_dataset}, followed by detailed descriptions in the subsequent paragraphs.

\begin{table}[htbp]
	\centering
	\caption{Overview of the five multimodal datasets utilized in our experiments, including modality composition, data scale, number of classes, and target task for each dataset.}
	\label{tab:appendix_dataset}
	\begin{adjustbox}{width=\linewidth}
		\begin{tabular}{lllll}
			\toprule
			\textbf{Dataset} & \textbf{Modality} & \textbf{Scale} & \textbf{Classes} & \textbf{Task} \\ 
			\midrule
			AV-MNIST & Image, Audio & 70k & 10 & Handwritten digit recognition \\
			RAVDESS (Speech) & Video, Audio & 1,440 & 8 & Emotion recognition \\
			VGGSound-50k & Video, Audio & 49k & 141 & Scene classification \\
			CrisisMMD-V2 & Image, Text & 16058 & 8 & Social media comment prediction \\
			NYU-Depth-V2 & RGB images, Depth images & 1,449 & 41 & Indoor scene semantic segmentation \\
			\bottomrule
		\end{tabular}
	\end{adjustbox}
\end{table}

\paragraph{AV-MNIST Dataset}
The AV-MNIST dataset is a multimodal benchmark for digit recognition integrating visual and auditory modalities, comprising 70,000 paired samples across 10 classes, where the visual modality consists of $28 \times 28$ MNIST images with 75\% of their energy removed via principal component analysis (PCA) to simulate low-quality input, while the auditory modality is derived from 25,102 spoken digit audio samples of the Tidigits database, augmented by overlaying noise segments from the ESC-50 dataset, and represented as $112 \times 112$ spectrograms.

\paragraph{RAVDESS Dataset}
The RAVDESS dataset is a multimodal benchmark for emotion recognition, comprising 1,440 audiovisual samples of emotional utterances. Both video and audio modalities are preprocessed to a uniform duration of 3.6 seconds, with visual inputs derived from 15 uniformly sampled frames, further processed using the  MTCNN to extract facial regions from video frames. The audio modality is represented by 15-dimensional Mel-frequency cepstral coefficients (MFCCs) extracted at a sampling rate of 22,050 Hz.

\paragraph{VGGSound-50k Dataset}
The VGGSound-50k dataset constitutes a high-quality multimodal benchmark comprising 48,755 YouTube video clips across 141 fine-grained audio-visual scene categories, derived from VGGSound-AVEL50k through rigorous quality filtering. Following AVEL's methodology, we extract visual features using VGG19 at 16fps and audio features via VGGish with standard preprocessing, maintaining temporal alignment between modalities. With standardized feature dimensions (512D visual, 128D audio) and 10-second average clip duration, this dataset provides a robust testbed for advancing research in multimodal scene understanding.

\paragraph{CrisisMMD-V2 Dataset}
The CrisisMMD-V2 dataset serves as a multimodal benchmark for humanitarian crisis classification. It contains 16,058 annotated Twitter posts, each comprising a paired image and text, spanning 8 distinct humanitarian categories. Each sample includes manually verified semantic alignment between the visual and textual modalities, supporting joint vision-language modeling. For feature extraction, we utilize a pretrained ResNet-50 (with the classification head removed) for images, and BERTweet-base for textual representations.

\paragraph{NYU-Depth-V2 Dataset}
The NYU-Depth-V2 dataset serves as a widely adopted benchmark for indoor scene understanding, comprising 1,449 precisely aligned RGB-depth image pairs captured using Microsoft Kinect sensors in diverse indoor environments (e.g., homes, offices, and stores). Each image is meticulously annotated with 40 semantic categories (e.g., walls, floors, furniture) and dense pixel-level labels, facilitating research in semantic segmentation, depth estimation, and 3D reconstruction tasks.

\subsection{Model Configurations Across Datasets}
\label{appendix:model}
The network architectures of the multimodal model and the two corresponding unimodal models for the five multimodal datasets discussed in the main text are summarized in Table~\ref{tab:appendix_model_cfg}. For more implementation details, please refer to our released code.

\begin{table*}[htbp]
	\caption{Model configurations for each dataset, including two unimodal models and one multimodal model. The suffix \texttt{net\_woHead} indicates that the classification head of the corresponding unimodal network is removed, and the symbol $\odot$ denotes the feature concatenation operation. Here, \texttt{I}, \texttt{V}, and \texttt{T} represent image, video, and audio modalities, respectively.}
	\label{tab:appendix_model_cfg}
	\centering
	\begin{adjustbox}{width=\linewidth}
		\begin{tabular}{llll}
			\toprule
			\textbf{Dataset}      & \textbf{Unimodal Model 1}                       & \textbf{Unimodal Model 2}                 & \textbf{Multimodal Model}                                                                     \\
			\midrule
			AV-MNIST     & I: LeNet5                               & A: ThreeLayerCNN-2D               & \begin{tabular}[c]{@{}l@{}}Inet\_woHead $\odot$ FiveLayerCNN-2D\_woHead\\ + MLP\end{tabular} \\
			RAVDESS      & V: ThreeLayerCNN-3D + MLP                 & A: ThreeLayerCNN-1D + MLP           & \begin{tabular}[c]{@{}l@{}}Vnet\_woHead $\odot$ Anet\_woHead\\ + MLP\end{tabular}            \\
			VGGSound-50k & V: ThreeLayerCNN-3D + MLP                 & A: ThreeLayerCNN-1D + MLP           & \begin{tabular}[c]{@{}l@{}}Vnet\_woHead $\odot$ Anet\_woHead\\ + MLP\end{tabular}            \\
			CrisisMMD-V2 & I: MLP                                  & T: MLP                            & Inet\_woHead $\odot$ Tnet\_woHead + MLP                                                      \\
			NYU-Depth-V2 & RGB: FuseNet (RGB-Encoder) + Decoder\_RGB & D: FuseNet(D-Encoder) + Decoder\_D & FuseNet                                                                              \\
			\bottomrule
		\end{tabular}
	\end{adjustbox}
\end{table*}

\section{Supplementary Materials for MST-Distill}
\label{appendix:mst_others}
This section presents supplementary materials that validate key design choices, clarify implementation details, and provide additional evidence for the effectiveness of the proposed \textit{MST-Distill} framework. 

\subsection{MST-Distill: Pseudocode}
\label{appendix:mst_pseudocode}
The overall procedure of MST-Distill, encompassing the three stages described in main text, is summarized in Algorithm~\ref{alg:mst}.
\begin{algorithm}[htbp]
	\caption{MST-Distill Framework for Cross-Modal Knowledge Distillation}
	\label{alg:mst}
	\begin{algorithmic}[1]
		\REQUIRE Multimodal dataset $\mathcal{D} = \{(x_1^{(s)}, \dots, x_M^{(s)}; y^{(s)})\}_{s=1}^{S}$;\\
		Target student modality index $t \in \{1,2,...,M\}$
		\ENSURE Trained student model $f_{m_t}$ with optimal parameters $\theta_{m_t}^*$
		
		\STATE \textbf{Stage 1: Collaborative Initialization (CI)}
		\STATE Initialize $\{ \theta_{{m_i}} \}_{i=0}^{M}$ for all modality-specific models $\{ f_{{m_i}} \}_{i=0}^{M}$
		\STATE $\{ \theta_{{m_i}}^* \}_{i=0}^{M} \gets \arg\min_{\{ \theta_{{m_i}} \}_{i=0}^{M}} \; \mathcal{L}_{\mathrm{S_{1}}}$ \COMMENT{Equation~\eqref{eq:s1}}
		
		\STATE \textbf{Stage 2: Specialized Teacher Adaptation (STA)}
		\STATE Load $\{ \theta_{{m_i}}^* \}_{i=0}^{M}$ for all models and initialize $\{ \theta_{\mathrm{MN}}^j \}_{j=1}^{N}$
		\STATE Identify target student modality $m_t$ and corresponding student model $f_{m_t}$
		\STATE Construct specialized teachers $\{ f_{m_{\delta(j)}}^j \}_{j=1}^{N}$ via MaskNet insertion
		\STATE Freeze all model parameters except $\{ \theta_{\mathrm{MN}}^j \}_{j=1}^{N}$
		\FOR{each specialized teacher $f_{m_{\delta(j)}}^j$}
		\STATE Apply soft masking via MaskNet during forward pass
		\STATE $\theta_{\mathrm{MN}}^{j*} \gets \arg\min_{\theta_{\mathrm{MN}}^j} \; \mathcal{L}_{\mathrm{S}_{2}^{j}}$ \COMMENT{Equation~\eqref{eq:s2}}
		\ENDFOR
		
		\STATE \textbf{Stage 3: Dynamic Knowledge Distillation (DKD)}
		\STATE Load $f_{m_t}$ with $\theta_{m_t}^*$ from Stage 1
		\STATE Load specialized teachers $\{ f_{m_{\delta(j)}}^j \}_{j=1}^{N}$ with base parameters $\{\theta_{m_{i}}^*\}_{i=0, i\ne k}^{M}$ from Stage 1 and MaskNet parameters $\{\theta_{\mathrm{MN}}^{j*}\}_{j=1}^{N}$ from Stage 2
		\STATE Initialize $\theta_{\mathrm{GN}}$ for GateNet
		\STATE Freeze all parameters except $\theta_{m_t}$ and $\theta_{\mathrm{GN}}$
		\FOR{each minibatch $B$ from $\mathcal{D}$}
		\STATE Compute student logits $Z_{\text{out}} = f_{m_t}(x_t; \theta_{m_t})$
		\STATE Select top-$k$ teachers based on GateNet confidence scores
		\STATE $\{\theta_{m_t}^*, \theta_{\mathrm{GN}}^*\} \gets \arg\min_{\theta_{m_t}, \theta_{\mathrm{GN}}} \; \mathcal{L}_{\mathrm{S_{3}}}$ \COMMENT{Equation~\eqref{eq:s3}}
		\ENDFOR
		\RETURN $f_{m_t}(\cdot;\theta_{m_t}^*)$
	\end{algorithmic}
\end{algorithm}

\subsection{Feature Visualization Analysis}
To further understand the working mechanisms of our proposed methods, we conduct comprehensive visualization analyses on different datasets and components.

\begin{figure*}[htbp]
	\centering
	\includegraphics[width=0.83\linewidth]{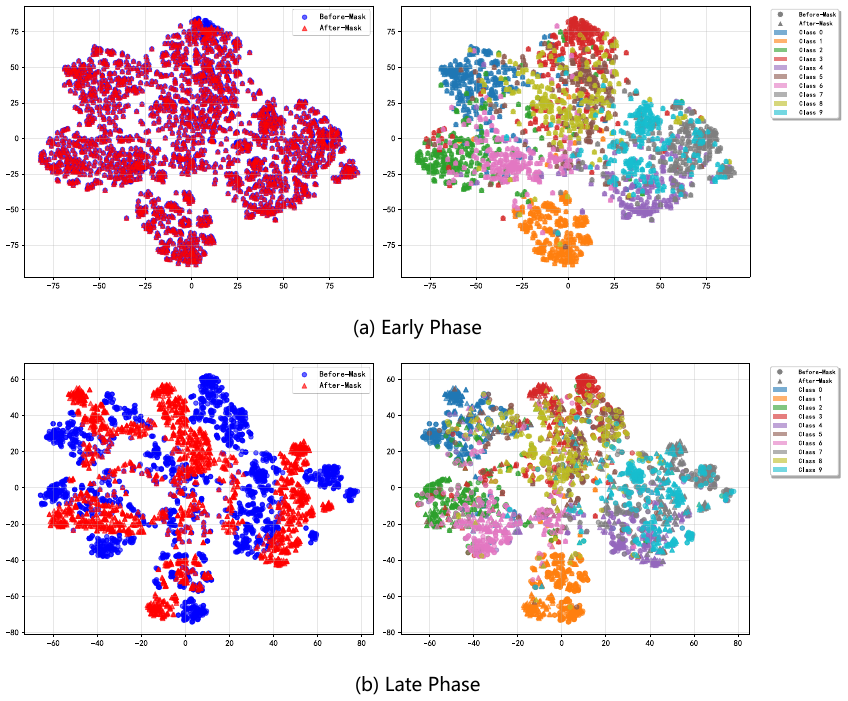}
	\caption{t-SNE visualizations of teacher features before and after MaskNet processing on AV-MNIST dataset. (a) Early phase and (b) Late phase of the STA stage. Left: features colored by processing status; Right: features colored by class labels.}
	\label{fig:appendix_vis_tsne}
\end{figure*}

For MaskNet, we perform t-SNE visualization on 2048 samples from the AV-MNIST validation set when the visual modality is selected as the target student modality. We extract features from the penultimate layer of the multimodal teacher before and after MaskNet processing during the early and late phases of the Specialized Teacher Adaptation (STA) stage, as shown in Figure~\ref{fig:appendix_vis_tsne}. The results reveal that in early training phase, MaskNet only affects partial samples within few categories (e.g., class 3). As training progresses, MaskNet's influence becomes more pronounced, exhibiting finer-grained instance-level reconstruction that leads to further intra-class cluster differentiation. This aligns with our intuition that modality differences exhibit sample-level distinctions, validating the module's effectiveness in suppressing inter-modal disparities.

\begin{figure*}[htbp]
	\centering
	\includegraphics[width=0.83\linewidth]{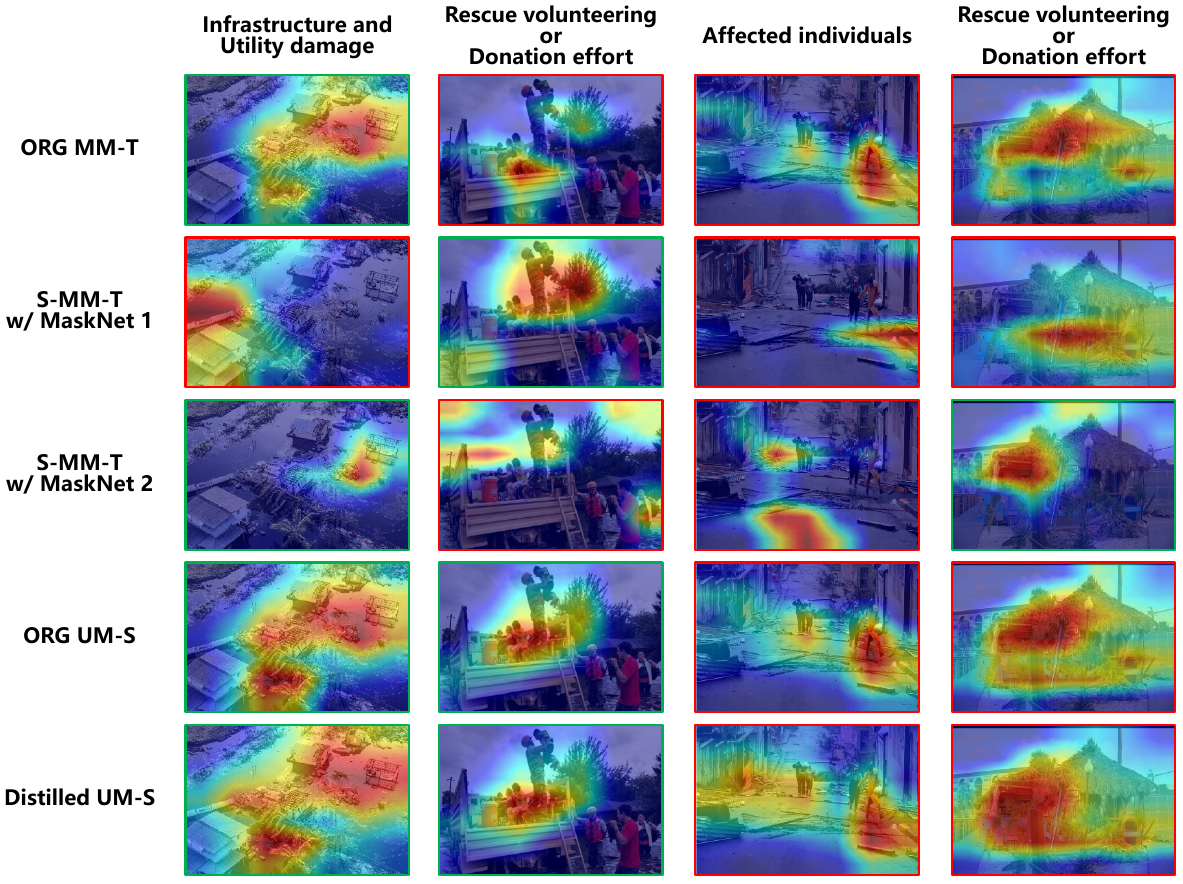}
	\caption{Grad-CAM visualizations of different models on four CrisisMMD-V2 samples. Each row shows the visualizations produced by a specific model, with the ground-truth label shown atop each column. Green and red borders indicate correct and incorrect predictions, respectively. MM-T and UM-S denote the multimodal teacher and unimodal student, respectively. The prefix \texttt{ORG} denotes models trained from scratch, while \texttt{S-} indicates the Specialized Teachers within the MST-Distill framework.}
	\label{fig:appendix_vis_heatmap}
\end{figure*}

For MST, we conduct Grad-CAM visualization on easy and hard samples from the CrisisMMD-V2 test set, as shown in Figure~\ref{fig:appendix_vis_heatmap}. Compared to multimodal teachers trained from scratch, the Specialized Teachers within the MST-Distill framework consistently extract more diverse visual cues after integrating different MaskNet variants. This enriches the teacher's representation and helps the distilled student model attend to broader informative regions than independently trained students.

\begin{table}[htbp]
	\centering
	\caption{Computational overhead comparison across different knowledge distillation methods on AV-MNIST dataset.}
	\begin{adjustbox}{width=\linewidth}
		\begin{tabular}{lcccccc}
			\toprule
			\textbf{Method} & KD & FitNets & RKD & MGDFR & C$^2$KD & MST-Distill \\
			\midrule
			\textbf{Peak Memory (GB)} & 9.1 & 13.2 & 13.6 & 7.6 & 7.1 & 53.6 \\
			\textbf{Training Time (s)}  & 442.3 & 1171.6 & 597.1 & 491.0 & 310.9 & 1942.2  \\
			\bottomrule
		\end{tabular}
		\label{tab:appendix_ce}
	\end{adjustbox}
\end{table}


\begin{table}[htbp]
	\centering
	\caption{Performance comparison between DML-style training and joint training across four multimodal datasets. The best performance in each dataset is highlighted in \textbf{bold}.}
	\begin{adjustbox}{width=\linewidth}
		\begin{tabular}{lcccc}
			\toprule
			\textbf{Setting} & \textbf{AV-MNIST} & \textbf{RAVDESS} & \textbf{VGGSound-50k} & \textbf{CMMD-V2} \\
			\midrule
			DML-Style       & 0.5363            & 0.7410           & 0.5326                & 0.5479           \\
			\textbf{Joint Training (Ours)}  & \textbf{0.5370}   & \textbf{0.7521}  & \textbf{0.5347}       & \textbf{0.5481}  \\
			\bottomrule
		\end{tabular}
		\label{tab:appendix_ci}
	\end{adjustbox}
\end{table}


\begin{table}[htbp]
	\centering
	\caption{Performance comparison of different load balancing losses across four multimodal datasets. CV and KL denote the Coefficient of Variation loss and the Kullback–Leibler divergence-based loss, respectively.}
	\begin{adjustbox}{width=\linewidth}
		\begin{tabular}{lcccc}
			\toprule
			\textbf{LB Loss} & \textbf{AV-MNIST} & \textbf{RAVDESS} & \textbf{VGGSound-50k} & \textbf{CMMD-V2} \\
			\midrule
			CV & \textbf{0.5370} & 0.7514 & 0.5326 & 0.5449 \\
			\textbf{KL (Ours)} & \textbf{0.5370} & \textbf{0.7521} & \textbf{0.5347} & \textbf{0.5481} \\
			\bottomrule
		\end{tabular}
		\label{tab:appendix_lb}
	\end{adjustbox}
\end{table}

\subsection{Computational Efficiency Analysis}
To evaluate computational overhead, we analyze our method using AV-MNIST as a representative benchmark, where computational patterns remain consistent across multimodal datasets. Training time measures effective iterations to convergence with early stopping, while peak memory indicates maximum GPU consumption. For single-teacher baselines, training time represents cumulative cost across multiple sessions due to separate teacher selection per modality.
As shown in Table~\ref{tab:appendix_ce}, MST-Distill requires 53.6 GB peak memory and 1942.2s training time, the highest among all methods. While this overhead stems from simultaneous multi-teacher processing and complex cross-modal knowledge transfer, it is justified by substantial performance gains. Since cross-modal distillation aims for high-performance student models with efficient inference, the one-time training cost is acceptable given persistent deployment benefits. Furthermore, MaskNet's parameter count can be flexibly adjusted during deployment by selecting teacher reconstruction layers and hidden nodes, enabling performance-cost trade-offs with considerable optimization potential.

\subsection{Collaborative Initialization Approaches}
We conduct experiments on the model collaborative initialization stage using two strategies: DML-style training with gradient detachment and joint training without it. The average performance of target unimodal student models across four multimodal classification datasets is summarized in Table~\ref{tab:appendix_ci}. It can be observed that modality-specific models trained via joint training achieve better performance within the MST-Distill framework. This suggests that gradient accumulation along shared optimization paths benefits subsequent knowledge transfer.

\subsection{Loading Balance Loss}
To validate the choice of load balancing (LB) loss in our MST-Distill framework, we compare the commonly used Coefficient of Variation (CV) loss with our Kullback–Leibler divergence-based design. While CV is widely adopted in Mixture-of-Experts (MoE) models to balance expert utilization, our KL loss is formulated based on Kullback–Leibler divergence to encourage the teacher routing distribution to approximate a uniform distribution. As shown in Table~\ref{tab:appendix_lb}, the KL-based loss yields consistently better performance, demonstrating its effectiveness in facilitating dynamic knowledge distillation.

\end{document}